\pdfoutput=1
\documentclass[conference]{IEEEtran}
\usepackage{times}
\usepackage[numbers]{natbib}
\usepackage{multicol}
\newcommand{\lmr}[1]{{#1}}
\usepackage[colorlinks,linkcolor=black,anchorcolor=black,urlcolor=blue]{hyperref}
\usepackage{amsmath,amsfonts,amssymb}
\usepackage{prettyref}
\usepackage{mathrsfs}
\usepackage{graphicx}
\usepackage{wrapfig}
\usepackage{subfig}
\usepackage{MnSymbol}
\usepackage{multirow}
\usepackage{booktabs}
\usepackage{algorithm}
\usepackage[noend]{algpseudocode}
\usepackage[table]{xcolor}
\usepackage{cellspace}
\usepackage{mathtools}

\newtheorem{theorem}{Theorem}[section]
\newtheorem{lemma}[theorem]{\TE{Lemma}}

\algnewcommand{\LineComment}[1]{\State \(\triangleright\) #1}
\algdef{SE}[DOWHILE]{Do}{doWhile}{\algorithmicdo}[1]{\algorithmicwhile\ #1}
\newcommand*{\colorboxed}{}
\def\colorboxed#1#{%
  \colorboxedAux{#1}%
}
\newcommand*{\colorboxedAux}[3]{%
  \begingroup
    \colorlet{cb@saved}{.}%
    \color#1{#2}%
    \boxed{%
      \color{cb@saved}%
      #3%
    }%
  \endgroup
}

\newrefformat{Fig}{Figure~\ref{#1}}
\newrefformat{fig}{Figure~\ref{#1}}
\newrefformat{par}{Section~\ref{#1}}
\newrefformat{appen}{Appendix~\ref{#1}}
\newrefformat{sec}{Section~\ref{#1}}
\newrefformat{sub}{Section~\ref{#1}}
\newrefformat{table}{Table~\ref{#1}}
\newrefformat{alg}{Algorithm~\ref{#1}}
\newrefformat{Alg}{Algorithm~\ref{#1}}
\newrefformat{Def}{Definition~\ref{#1}}
\newrefformat{Thm}{Theorem~\ref{#1}}
\newrefformat{Lem}{Lemma~\ref{#1}}
\newrefformat{step}{Step~\ref{#1}}
\newrefformat{ln}{Line~\ref{#1}}
\newrefformat{eq}{Equation~\ref{#1}}
\newrefformat{pb}{Problem~\ref{#1}}
\newrefformat{it}{Item~\ref{#1}}
\newrefformat{te}{Term~\ref{#1}}
\def\Eqref Eq:#1:{\eqref{eq:#1}}
\newrefformat{Eq}{Equation~\Eqref#1:}

\newcommand{\E}[1]{\mathbf{#1}}
\newcommand{\TE}[1]{\textbf{#1}}

\newcommand{\FPPROW}[2]{{\partial{#1}}/{\partial{#2}}}

\newcommand{\FPP}[2]{\frac{\partial{#1}}{\partial{#2}}}

\newcommand{\FPPTT}[3]{\frac{\partial^2{#1}}{\partial{#2}\partial{#3}}}

\newcommand{\TWO}[2]{\left(\setlength{\arraycolsep}{1pt}\begin{array}{cc}{#1}, & {#2}\end{array}\right)}
\newcommand{\TWOC}[2]{\left(\setlength{\arraycolsep}{1pt}\begin{array}{c}#1 \\ #2\end{array}\right)}
\newcommand{\TWOR}[2]{\left(\setlength{\arraycolsep}{1pt}\begin{array}{cc}{#1}^T, & {#2}^T\end{array}\right)^T}
\newcommand{\THREE}[3]{\left(\setlength{\arraycolsep}{1pt}\begin{array}{ccc}{#1}, & {#2}, & {#3}\end{array}\right)}

\newcommand{\THREEC}[3]{\left(\setlength{\arraycolsep}{1pt}\begin{array}{c}#1 \\ #2 \\ #3\end{array}\right)}

\newcommand{\FOURC}[4]{\left(\setlength{\arraycolsep}{1pt}\begin{array}{c}#1 \\ #2 \\ #3 \\ #4\end{array}\right)}

\newcommand{\EIGHTC}[8]{\left(\setlength{\arraycolsep}{1pt}\begin{array}{c}#1 \\ #2 \\ #3 \\ #4 \\ #5 \\ #6 \\ #7 \\ #8\end{array}\right)}

\newcommand{\MTT}[4]{\left(\setlength{\arraycolsep}{1pt}\begin{array}{cc}#1 & #2 \\ #3 & #4\end{array}\right)}

\newcommand{\MDD}[3]{\left(\setlength{\arraycolsep}{1pt}\begin{array}{ccc}#1 & & \\ & #2 & \\ & & #3\end{array}\right)}

\newcommand{\fmin}[1]{\underset{#1}{\E{min}}\;}
\newcommand{\fmax}[1]{\underset{#1}{\E{max}}\;}
\newcommand{\argmin}[1]{\underset{#1}{\E{argmin}}\;}
\newcommand{\argminP}[1]{\E{argmin}\;}
\newcommand{\argmax}[1]{\underset{#1}{\E{argmax}}\;}
\newcommand{\argmaxP}[1]{\E{argmax}\;}
\newcommand{\ST}{\E{s.t.}\;}

\definecolor{darkgreen}{HTML}{186a3b}
\newcommand{\LOSS}{\mathcal{L}}
\newcommand{\TRI}{\mathcal{T}}
\newcommand{\PT}{\E{p}}
\newcommand{\DISTGRASP}{\mathcal{L}_{data}}
\newcommand{\DIST}{\E{d}}
\newcommand{\NOR}{\E{n}}
\newcommand{\WR}{\E{w}}
\newcommand{\FF}{\E{f}}
\newcommand{\TT}{\boldsymbol{\tau}}
\newcommand{\HULL}{\mathbb{H}}
\newcommand{\RRR}{\mathbb{R}}
\newcommand{\WWW}{\mathbb{W}}
\newcommand{\METRIC}{\mathbb{M}}
\newcommand{\EEXP}{\E{exp}}
\newcommand{\DD}{\E{s}}

\newcommand{\VV}{\E{v}}
\newcommand{\Id}{\E{I}}
\newcommand{\TAN}{\E{t}}
\newcommand{\SDPBASIS}{\mathcal{V}}
\newcommand{\NEURAL}{\mathcal{N}}
\newcommand{\COEF}{c}
\newcommand{\EDGE}{\E{e}}
\newcommand{\VERTEX}{\E{v}}
\newcommand{\CC}{\E{c}}

\newcommand{\XX}{\E{x}}
\newcommand{\ZZZ}{\E{Z}}
\newcommand{\ZZZZ}{\mathcal{Z}}
\newcommand{\FFF}{\E{F}}
\newcommand{\FFFF}{\mathcal{F}}
\newcommand{\SVEC}[1]{\E{svec}(#1)}

\usepackage{sidecap}

\IEEEoverridecommandlockouts
\begin{document}
\captionsetup[figure]{font={small}}
\title{\scalebox{0.9}{Deep Differentiable Grasp Planner for High-DOF Grippers}\vspace{-20px}}

\newif\ifreview
\ifreview
\author{Author Names Omitted for Anonymous Review. Paper-ID [62]}
\else
\author{Min Liu$^{1,3}$, Zherong Pan$^{2}$, Kai Xu$^{1*}$, Kanishka Ganguly$^{3}$, and Dinesh Manocha$^{3}$  \\
{$^{1}$School of Computer, National University of Defense Technology}\\
{$^{2}$Department of Computer Science, University of North Carolina at Chapel Hill}\\
{$^{3}$Department of Computer Science and Electrical \& Computer Engineering, University of Maryland at College Park}\\
{\href{https://gamma.umd.edu/researchdirections/grasping/differentiable_grasp_planner}{https://gamma.umd.edu/researchdirections/grasping/differentiable\_grasp\_planner}}
\thanks{$^{*}$Correspondence to kevin.kai.xu@gmail.com.}\vspace{-12px}}


\maketitle
\setlength\abovedisplayskip{2pt}
\setlength\belowdisplayskip{2pt}
\begin{abstract}
We present an end-to-end algorithm for training deep neural networks to grasp novel objects. Our algorithm builds all the essential components of a grasping system using a forward-backward automatic differentiation approach, including the forward kinematics of the gripper, the collision between the gripper and the target object, and the metric for grasp poses. In particular, we show that a generalized $Q_1$ grasp metric is defined and differentiable for inexact grasps generated by a neural network, and the derivatives of our generalized $Q_1$ metric can be computed from a sensitivity analysis of the induced optimization problem. We show that the derivatives of the (self-)collision terms can be efficiently computed from a watertight triangle mesh of low-quality. Altogether, our algorithm allows for the computation of grasp poses for high-DOF grippers in an unsupervised mode with no ground truth data, or it improves the results in a supervised mode using a small dataset. Our new learning algorithm significantly simplifies the data preparation for learning-based grasping systems and leads to higher qualities of learned grasps on common 3D shape datasets \cite{calli2015benchmarking,singh2014bigbird,kasper2012kit,2015_ICRA_kbs}, achieving a $22\%$ higher success rate on physical hardware and a $0.12$ higher value on the $Q_1$ grasp quality metric.
\end{abstract}
\IEEEpeerreviewmaketitle
\section{\label{sec:intro}Introduction}
Robot grasping of unknown objects is an important problem and an essential component of various applications, including robot object packing \cite{wang2019stable,Hauser-RSS-19} and dexterous manipulation \cite{897777,doi:10.1177/0278364919872532}. Earlier methods \cite{6499286,ciocarlie2007dexterous,miller2000graspit,Dai2018} could generate grasp poses for an arbitrary gripper or target object, but they ignored the uncertainty of real world situations. Recent learning-based methods \cite{inproceedingsDexNetTwo,8460609,8460875,8463204,Mousavian_2019_ICCV,8304630,liu2019grasp} have demonstrated improved robustness in terms of handling sensor noise. Instead of directly inferring the grasp poses, these methods propose learning various intermediary information such as grasp quality measures \cite{inproceedingsDexNetTwo} or reconstructed 3D object shapes \cite{8460609} and then use this information to help infer grasp poses. On the positive side, it has been shown that learning this kind of information can improve both the data-efficacy of training and the success rate of predicted grasp poses. On the negative side, however, this intermediary information complicates the training procedure, hyper-parameter search, and data preparation \cite{8460609}.

\begin{figure}[ht]
\centering
\includegraphics[width=0.4\textwidth]{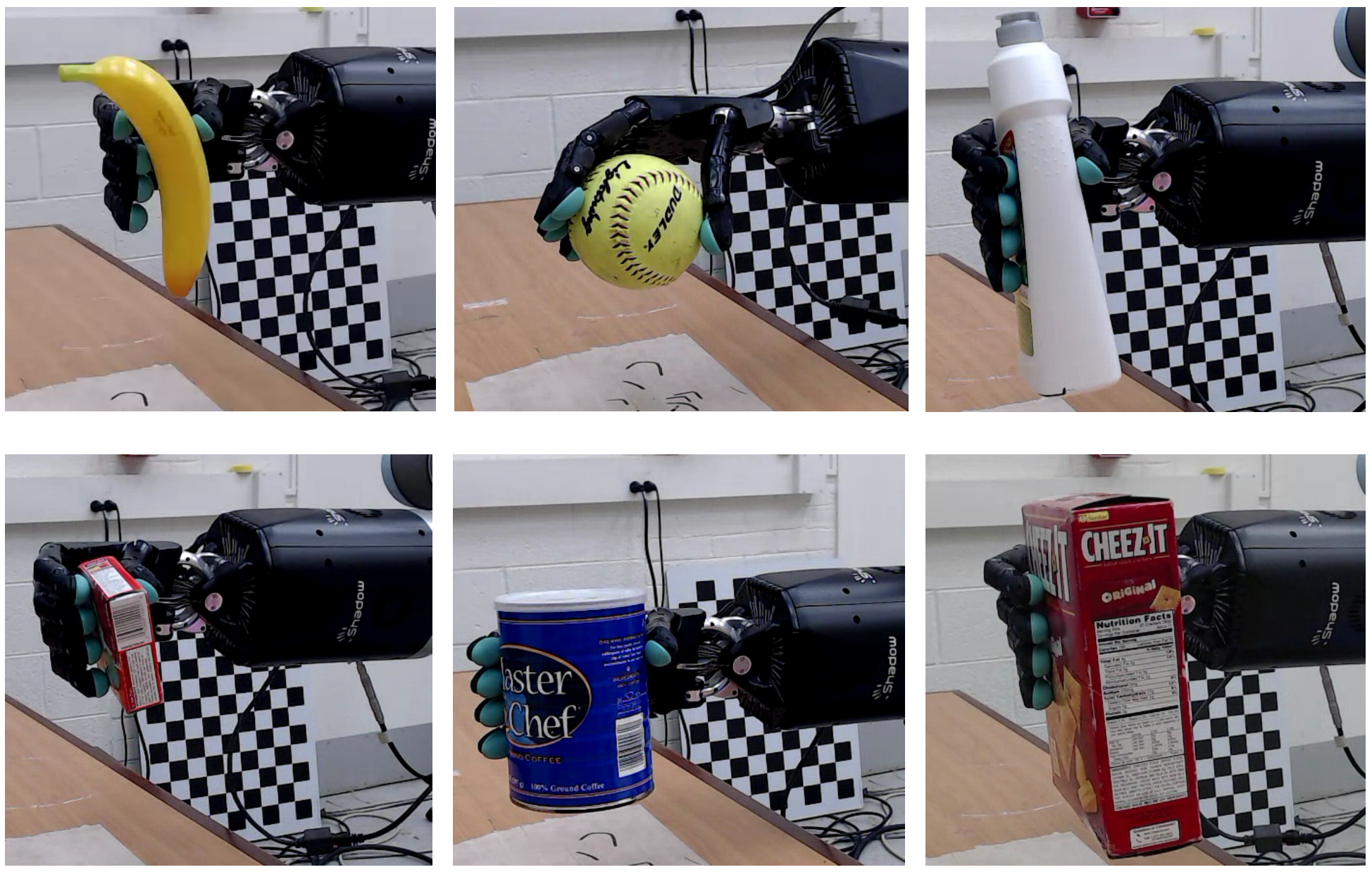}
\caption{\label{fig:results} Using a small dataset, we train an end-to-end neural network to predict grasp poses for novel objects that it has never see before. The neural network prediction is adjusted using our differentiable grasp quality metric.}
\vspace{-20px}
\end{figure}
Ideally, a learning-based grasp planner should infer the grasp poses directly from raw sensor inputs such as RGB-D images. Such approaches have been developed by many researchers \cite{saxena2008robotic,jiang2011efficient}. However, recent methods \cite{inproceedingsDexNetTwo,8460875} show that it is preferable to first learn a grasp quality metric function and then optimize the metric at runtime for an unknown target object using sampling-based optimization algorithms, such as multi-armed bandits \cite{mahler2016dex}. Such optimization can be very efficient for low-DOF parallel jaw grippers but less efficient for high-DOF anthropomorphic grippers due to their high-dimensional configuration spaces. In addition, it is possible for the sampling algorithm to generate samples at any point in the configuration space, and the learned metric function has to return accurate values for all these samples. To achieve high accuracy, a large amount of training data is needed, as shown in the 6.7 million ground truth grasps in the dataset used by \cite{inproceedingsDexNetTwo}.

Various techniques have been proposed to improve the robustness and efficiency of grasp planner training. Prior works \cite{8304630,8460609} proposed improving the data-efficiency of training by having the neural network recover the 3D volumetric representation of the target object from 2D observations. A 2D-to-3D reconstruction sub-task allows the model to learn intrinsic features about the object. However, a volumetric representation also incurs higher computational and memory cost. In addition, compared with surface meshes, volumetric representations based on signed distance fields cannot resolve delicate, thin features of complex objects \cite{liu2019grasp}. Demonstrating an alternative method, prior works in \cite{fang2018mtda,Mousavian_2019_ICCV} show that higher robustness can also be achieved using adversarial training, which in turn introduces additional sub-tasks of training and requires new data.

\begin{figure*}[ht]
\centering
\includegraphics[width=0.8\textwidth]{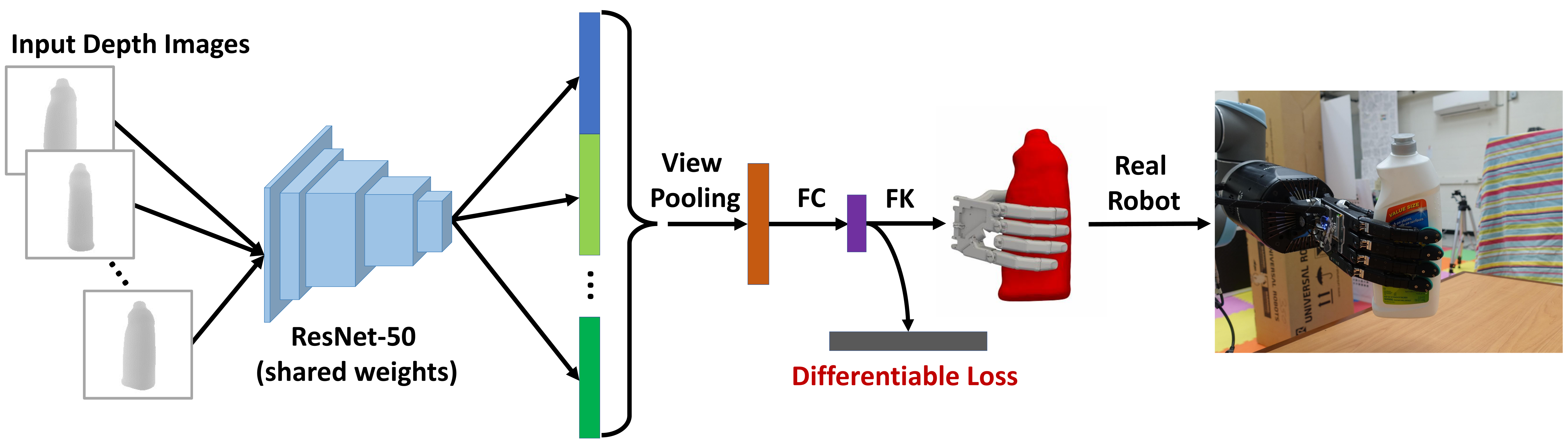}
\caption{\label{fig:architecture} Our learning architecture takes multi-view depth images of the object as inputs. The features of these images are extracted using ResNet-50, and these are then fed into the fully connected (FC) blocks after view pooling \cite{su2015multi} to predict the high-DOF configuration of a gripper directly. The configuration space is then brought through a forward kinematics (FK) block and transformed into Euclidean space. We then execute grasps of these configurations in a physical platform. During the training stage, we can formulate various requirements for a grasp planner as loss functions in Euclidean space (red), including (self-)collision-free, grasp quality maximization, data consistency, and closeness between the gripper and the target object's surface. Our method can be used as a locally optimal grasp planner guided by analytic gradients, or as an additional loss function to improve the quality of learned grasp poses.}
\vspace{-15px}
\end{figure*}
\TE{Main Results:} We present a differentiable theory of grasp planning, extending ideas from \cite{chen1993finding}, an early attempt to formulate grasp planning as a continuous optimization. Our main contribution is a generalized definition of the grasp quality metric that is defined when the gripper is not in contact with the target object. We show that this metric function is locally differentiable and that its gradient can be computed from the sensitivity analysis of the optimality condition in a similar manner to \cite{amos2017optnet}. We also propose a loss function to ensure that grasps are (self-)collision-free in a differentiable manner, which can be computed from only surface meshes of target objects.

Our method can be used as a locally optimal grasp planner similar to simulated annealing \cite{miller2000graspit}, but our method is guided by analytic gradients and can quickly find a locally optimal solution. More importantly, our method can be used to improve the quality of learned grasp poses using a simple neural network architecture. Specifically, we use a network that takes as input a set of multi-view depth images of the target object and directly predicts a grasp pose for a high-DOF gripper. This design choice is preferable to those in prior work \cite{mahler2016dex} because it leads to a higher performance during runtime, as there is no need to optimize a learned grasp metric and we can obtain the grasp pose by a single forward propagation through the neural network.

By adding our differentiable loss, we show that the simple neural network architecture can predict high-quality grasps for the Shadow Hand (\prettyref{fig:results}) after training on a dataset of only 400 objects and 40K ground truth grasps. When compared with the supervised learning baseline \cite{liu2019grasp}, our method achieves a $22\%$ higher success rate on physical hardware and a $0.12$ higher value in the $Q_1$ grasp quality metric \cite{219918}. Our learning architecture is illustrated in \prettyref{fig:architecture}.

\section{\label{sec:related}Related Work}
In this section, we review related works in grasp planning that uses either model-based or learning-based methods.

\TE{Model-Based Grasp Planners} assume perfect sensing of environment geometries and target object shapes. Given the geometric information, a grasp planner searches for a grasp pose that maximizes a certain grasp quality metric; many techniques have been proposed for defining reasonable grasp quality metrics \cite{6335488,219918,schulman2017grasping,roa2015grasp} and designing efficient search algorithms \cite{Dai2018,219918,ciocarlie2007dexterous,miller2000graspit}. These methods can be applied to both low- and high-DOF grippers and can be classified into discrete sampling-based techniques \cite{miller2000graspit} and continuous optimization techniques \cite{chen1993finding}. Sampling-based methods allow virtually any grasp quality metric to be used as the objective function, while continuous methods require the metric to be differentiable with respect to the configuration of the gripper. In practice, continuous optimization techniques are more efficient in terms of finding the (locally) optimal grasp poses. \lmr{Some works \cite{kiatos2019grasping,maldonado2010robotic} plan grasps by optimizing differentiable losses. However, these losses do not directly measure grasp qualities.}

Some planning methods \cite{219918,7539662,7989253,hang2017a} only compute optimal grasp points, while others \cite{Dai2018,1909.05430} compute both the grasp points and the gripper poses. When a gripper pose is needed, the planner uses a two-stage approach: a set of grasp points is first selected on the surface of the target object and then the pose of the gripper is found by inverse kinematics. Based on the idea of numerically optimizing the grasp quality metric, we extend the definition of a grasp quality metric to be well-defined in the ambient space, i.e. when the gripper is not in contact with the target object, thereby unifying grasp points selection and gripper pose computation.

\TE{Learning-Based Grasp Planners} can predict grasp points or gripper poses given noisy observations of the environment. Most early works \cite{saxena2008robotic,saxena2007robotic,gualtieri2016high} in this direction assume that a parallel-jaw gripper is designed for the target object and that the input is a single depth image of the target object. In this case, the grasp problem boils down to selecting the gripper's initial direction and orientation, which can be solved using an analytic method \cite{jiang2011efficient}. A noteworthy success in this problem is achieved by DexNet \cite{mahler2016dex,inproceedingsDexNetTwo}, which uses deep convolutional neural networks to learn object similarity functions and grasp quality functions. DexNet can robustly pick a large number of unknown objects using a dataset of tens of thousands of target objects and millions of ground truth grasp poses.

More recent techniques aim to improve the data efficiency of learning-based planners and also make the planner robust in challenging settings involving \lmr{high-dimensional visual observation of the environment \cite{morrison2018closing}}, arbitrary approaching directions \cite{fang2018mtda}, more general gripper types \cite{8304630,1909.05430}, and model discrepancies \cite{james2019sim,tobin2017domain}. It has been shown in \cite{fang2018mtda}, among others, that the grasp planning task can be divided into two sub-tasks, object reconstruction and gripper pose prediction, and that learning these two sub-tasks can improve the rate of success. It is shown in \cite{8460609} that adversarial training can also improve the robustness of the learned model. However, these methods either perform extensive data generation or require delicate parameter tuning for the adversarial training.

A common drawback of prior works \cite{mahler2016dex,inproceedingsDexNetTwo,fang2018mtda,lu2020multifingered,lu2020planning,van2019learning} is that they learn a grasp quality metric function or grasping success predictor, which requires an additional sampling-based optimizer to search for gripper poses. This requirement limits these methods to low-DOF grippers, since the high-dimensional configuration space of high-DOF grippers makes the sampling-based optimization computationally costly. Some recent methods \cite{liu2019grasp} overcame this difficulty by directly predicting a nominal gripper pose from an observation of the object. However, the predicted gripper pose is not directly usable and needs to be post-processed. In comparison, our method predicts robust, usable gripper poses using a simple neural-network architecture and uses a smaller dataset for training. In addition, our method can be combined with previous learning-based methods to improve their results.

As an alternative to supervised learning, reinforcement learning allows a learned grasp planner to discover useful grasp poses through exploration. Learned grasp planners have been successfully applied to grasping \cite{quillen2018deep} and other manipulation problems \cite{zhu2019dexterous}. However, the number of state transition data needed in a typical training is on the level of millions \cite{quillen2018deep}, while we show that robust gripper poses can be predicted by supervised learning on a dataset with 400 example objects using 40K ground truth grasp poses.

\section{\label{sec:problem}Learning Grasp Poses for High-DOF Grippers}
Our goal is to learn a grasp prediction network $\NEURAL(D_{1,\cdots,K};\theta)$ from multiple depth images of the target object, where $D_i$ is the depth image taken from the $i$th camera view facing the target object to be grasped and $\theta$ are the learnable parameters. The output of $\NEURAL$ is both the 6D extrinsic parameters and $I$ joint angles of the gripper, i.e. $\NEURAL(\bullet)\in\RRR^{6+I}$. This is in contrast to prior works \cite{fang2018mtda,inproceedingsDexNetTwo}, where another grasp quality metric function or grasp successful predicate function $\bar{\NEURAL}(D_{1,\cdots,K},\bullet;\theta')$ is learned and $\bullet$ is a candidate grasp pose. Next, the grasp pose is found by maximizing $\bar{\NEURAL}$ at runtime using sampling-based algorithms such as multi-arm bandits \cite{mahler2016dex}. 

However, when the gripper is high-DOF, the maximization of $\bar{\NEURAL}$ becomes a search in a high-DOF configuration space, which is time-consuming. As a result, we choose to learn $\NEURAL$ instead of $\bar{\NEURAL}$. The major challenge in learning $\NEURAL$ is to resolve the ambiguity in grasp poses, because infinitely many grasp poses can have the same grasp quality for a target object but our neural network $\NEURAL$ can only predict one pose. In order to resolve this ambiguity in grasp poses, we need the dataset to be consistent. A consistent grasp pose dataset is one where all the ground truth grasp poses can be represented by a single neural network. To enforce consistency, one prior work \cite{liu2019grasp} attempted to train $\NEURAL$ by precomputing multiple grasp poses for each target object and used a Chamfer loss to have $\NEURAL$ pick the most consistent pose. However, the learned gripper poses cannot be used directly due to their low quality, and post-processing is needed to deploy the learned poses on physical hardware. 

We aim to further improve the quality of the learned function $\NEURAL$ without increasing the complexity of training in terms of either the amount of data or the network architecture. Instead, we are inspired by earlier works \cite{chen1993finding,219918,Dai2018}, which formulate grasp planning as a continuous optimization. We incorporate all the criteria of good grasps as additional loss functions in terms of stochastic optimization. It has recently been shown that gradients can be brought through complex numerical algorithms to provide additional guidance. These domain-specific differentiable models \cite{amos2017optnet,hu2019difftaichi,hu2019chainqueen} can significantly improve the convergence rate of neural-network training and reduce the amount of data needed. However, we need to overcome several difficulties when using these approaches for grasp planning:
\begin{itemize}
    \item All the existing grasp quality metrics have discontinuities \cite{zheng2012efficient}, so we have to modify them for differentiability.
    \item A grasp quality metric is only defined when the gripper and the target object have exact contact, which is generally not the case when gripper poses are being stochastically updated by the training algorithm.
    \item Our differentiable loss function is defined for a target object represented using triangle meshes. These triangle meshes come from well-known 3D shape datasets \cite{calli2015benchmarking,singh2014bigbird,kasper2012kit,2015_ICRA_kbs}, some of which are of low quality. If gradient computation becomes unreliable on low-quality meshes (with nearly degenerate triangles), training will be misled.
\end{itemize}
We present our design of loss functions and discuss how to address the three challenging problems in the next section.
\section{\label{sec:method}Differentiable Grasp Planner}
Our loss function $\LOSS$ is comprised of three terms: $\E{e}^{-Q_1}$, $\LOSS_{guide}$, and $\LOSS_{coll,self}$. The first term is a generalized $Q_1$ grasp metric \cite{219918} that measures the quality of a grasp using physics-based rules. However, when force closure is not satisfied, both the metric value and its gradients are zero. In this degenerate case, we add a second, heuristic term $\LOSS_{guide}$ that always provides a non-vanishing gradient. Our third term $\LOSS_{coll,self}$ penalizes both self-collision and collisions between the gripper and the target object.
\subsection{Notation}
Throughout the paper, we assume that a target object is defined by a watertight triangle mesh $\TRI$. As illustrated in \prettyref{fig:illusQ}, given a point $\PT$ in the workspace, we can also define the signed distance to $\TRI$ as $\DIST(\PT)$ and the outward normal with respect to $\TRI$ as $\NOR(\PT)$. In addition, we also define $\NOR_g(\PT)$ as the gripper normal, i.e. the outward normal direction on the gripper mesh. We further assume that the target object's center-of-mass coincides with the origin of the Cartesian coordinates. During grasping, the object will be under an external wrench $\WR=\TWOR{\FF}{\TT}$.

\begin{figure}[ht]
\vspace{-5px}
\centering
\includegraphics[width=0.3\textwidth]{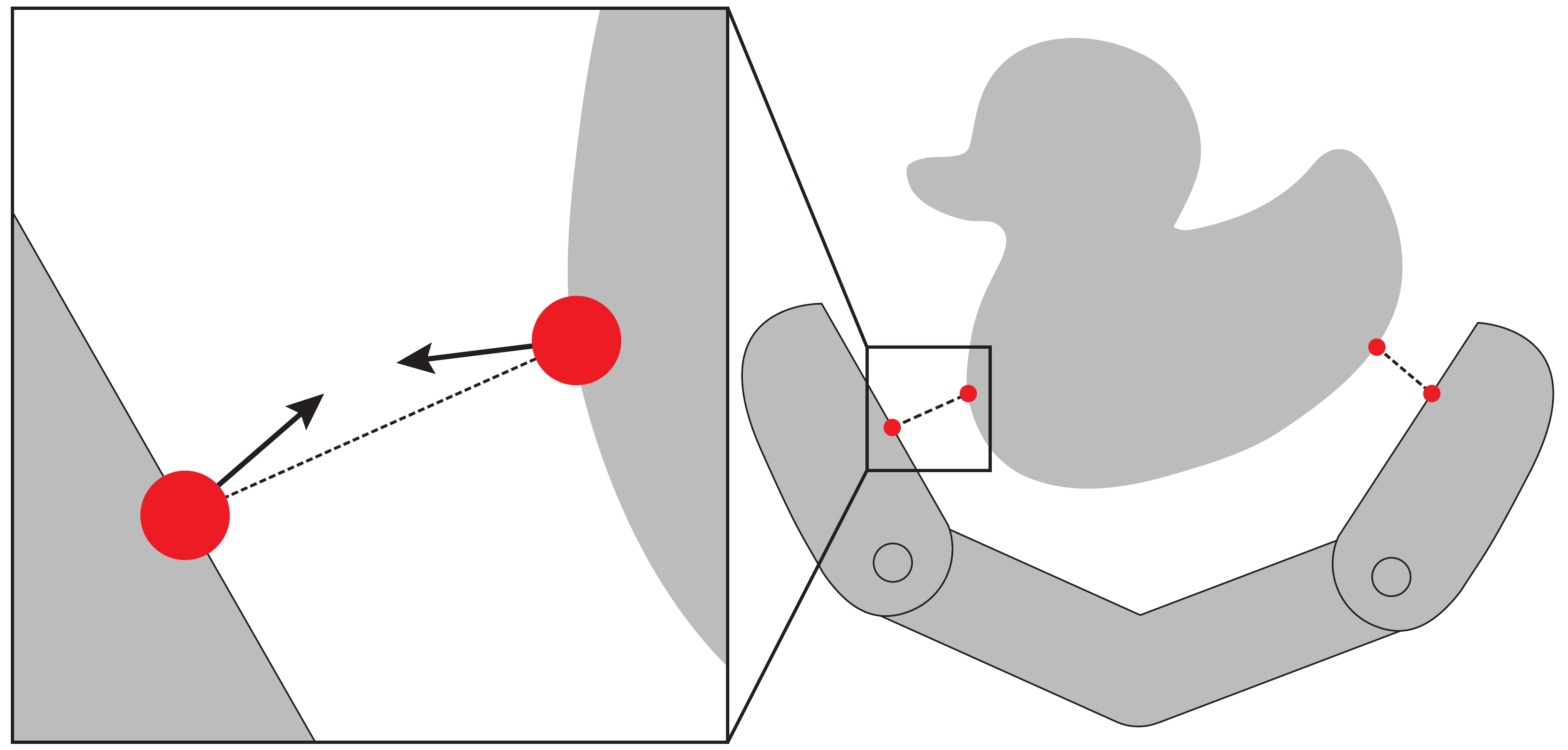}
\put(-125,20){\small\rotatebox{27}{$\DIST(\PT)$}}
\put(-117,45){\small\rotatebox{10}{$\NOR(\PT)$}}
\put(-143,29){\small\rotatebox{45}{$\NOR_g(\PT)$}}
\caption{\label{fig:illusQ} Variables used to define our generalized $Q_1$ metric.}
\vspace{-5px}
\end{figure}
For a set of grasp points $\PT_{1,\cdots,N}$ satisfying $\DIST(\PT_i)=0$, with respective grasping forces $\FF_i$, the quality of a grasp pose is defined by the $Q_1$ metric \cite{219918} as follows:
\begin{align}
\label{eq:Q1}
Q_1=&\fmax{\FF_i}r\quad\ST\{\WR|\|\sqrt{\METRIC}\WR\|\leq r\}\subseteq\WWW  \\
\WWW=&\{\sum_i\TWOC{\FF_i}{\PT_i\times\FF_i}
|
\TWOC{\sum_i\FF_i^T\NOR(\PT_i)\leq 1}
{\|\FF_i-\NOR(\PT_i)\NOR(\PT_i)^T\FF_i\|_2\leq\FF_i^T\NOR(\PT_i)\mu}\},\nonumber
\end{align}
where $\mu$ is the frictional coefficient and $\METRIC$ is the user-provided metric tensor that is equal to $\E{Diag}(1,1,1,m,m,m)$. \lmr{The metric $\METRIC$ is a standard way to tune the relative weights of force and torque. In our benchmarks, we simply choose $m$ to be inversely proportional to the object's average scale.} Intuitively, $Q_1$ is the radius of the origin-centered 6D sphere in the admissible wrench space, where an admissible wrench should satisfy two conditions: limited force magnitude and frictional cone constraints.

\subsection{Generalized $Q_1$ Metric with Inexact Contacts}
In practice, it is infeasible to assume that a grasp metric can be computed in its original form, i.e. \prettyref{eq:Q1}. This is because a learning system will generally not produce grasping points that lie exactly on the surface of the target object. It is well known that incorporating hard constraints into neural networks is difficult \cite{ravi2019explicitly}. When a stochastic training scheme is used and neural network parameters are randomly perturbed, exact constraint satisfaction will be lost. As a result, we have to deal with cases where $\DIST(\PT_i)\neq0$. Taking these cases into account, we derive a generalized version of $Q_1$ by modifying the first condition of admissible wrenches in \prettyref{eq:Q1} as follows:
\begin{equation}
\label{eq:Q1EXT}
\resizebox{.9\hsize}{!}{$
\sum_i\FF_i^T\NOR(\PT_i)
\EEXP(\alpha\|\DIST(\PT_i)\|+\beta(1+\NOR(\PT_i)^T\NOR_g(\PT_i)))\leq 1,
$}
\end{equation}
which essentially extends $Q_1$ to the ambient space by an exponential weight function with two terms. The first term $\|\DIST(\PT_i)\|$ ensures that our generalized $Q_1$ attains larger values when grasp points are closer to the surface of the target object. The second term $\beta(1+\NOR(\PT_i)^T\NOR_g(\PT_i))$ ensures that our generalized $Q_1$ attains larger values when the normal direction on the gripper and the normal direction on the target object align. Finally, it is obvious that \prettyref{eq:Q1EXT} converges to \prettyref{eq:Q1} as $\alpha,\beta\to\infty$. Like previous works \cite{tassa2012synthesis,mordatch2012discovery} on generalized contact-implicit models, our generalized metric allows a learning algorithm to determine the number of contact points and their positions. 

To train neural networks using the generalized $Q_1$ metric, we need to compute its sub-gradient with respect to $\PT_i$ efficiently. Unfortunately, the exact computation of the $Q_1$ metric is difficult because the optimization in \prettyref{eq:Q1} is non-convex; several approximations have been proposed in \cite{schulman2017grasping,Dai2018,zheng2012efficient}. We present two different techniques for computing $Q_1$ and $\partial{Q_1}/\partial{\PT_i}$. The first method computes an upper bound of generalized $Q_1$, which is cheaper to compute but creates zero entries in the gradient vector. The second method computes a smooth, lower bound of generalized $Q_1$, which propagates non-zero gradient information but costlier to compute.

\subsubsection{Derivatives of the $Q_1$ Upper Bound}
Our first technique adopts \cite{schulman2017grasping}, which approximates $Q_1$ by assuming that $\WR$ must be along one of a discrete set of directions: $\DD_{1,\cdots,D}$. This assumption results in a tractable upper bound of $Q_1$ and can be extend to our generalized $Q_1$ metric as follows:
\begin{equation}
\begin{aligned}
\label{eq:Q1U}
Q_1=&\fmin{j=1,\cdots,D}\left[\fmax{\WR\in\WWW}\DD_j^T\METRIC\WR\right],
\end{aligned}
\end{equation}
which is a min-max optimization. Here the minimization is with respect to a set of discrete indices, for which sub-gradients can be computed. The maximization aims at finding the support of $\DD_j$ in $\WWW$, and its optimal solution can be derived in a closed form. To show this, we first define the convex wrench space of each contact point $\PT_i$ as:
\begin{equation*}
\resizebox{.95\hsize}{!}{$
\WWW_i\triangleq\{\TWOC{\FF_i}{\PT_i\times\FF_i}|
\TWOC{\FF_i^T\NOR(\PT_i)\leq \EEXP(-\alpha\|\DIST(\PT_i)\|-\beta(1+\NOR(\PT_i)^T\NOR_g(\PT_i)))}
{\|\FF_i-\NOR(\PT_i)\NOR(\PT_i)^T\FF_i\|_2\leq\FF_i^T\NOR(\PT_i)\mu}\}.
$}
\end{equation*}
Then it is easy to verify that $\WWW=\E{ConvexHull}(\WWW_{1,\cdots,N})$ and the support of union of convex hulls is the maximum support of each hull, i.e.:
\begin{small}
\begin{align*}
\fmax{\WR\in\WWW}\DD_j^T\METRIC\WR=\fmax{i=1,\cdots,N}\fmax{\WR\in\WWW_i}\DD_j^T\METRIC\WR.
\end{align*}
Finally, the support of $\DD_j$ in $\WWW_i$ can be computed analytically as follows:
\begin{align*}
\fmax{\WR\in\WWW_i}\DD_j^T\WR=&
\EEXP(-\alpha\|\DIST(\PT_i)\|-\beta(1+\NOR(\PT_i)^T\NOR_g(\PT_i))) \\
&\begin{cases*}
\WR_\perp+\frac{\WR_\parallel^2}{\WR_\perp} & if $\mu\WR_\perp>\WR_\parallel$   \\
\fmax{}(0,\WR_\perp+\mu\WR_\parallel) & otherwise
\end{cases*}    \\
\WR_\perp\triangleq&\DD_j^T\METRIC\TWOC{\Id}{\PT_i\times}\NOR(\PT_i)    \\
\WR_\parallel\triangleq&
\left\|\DD_j^T\METRIC\TWOC{\Id}{\PT_i\times}\left[\Id-\NOR(\PT_i)\NOR(\PT_i)^T\right]\right\|.
\end{align*}
\end{small}
In this form, each operation for computing our generalized $Q_1$ can be implemented as a standard math operation with derivatives that can be computed using automatic differentiation tools such as \cite{paszke2017automatic}.

\subsubsection{Derivatives of the $Q_1$ Lower Bound \label{sec:KKT}}
We have shown that computing an upper bound of $Q_1$ reduces to a series of simple operations with well-defined sub-gradients. However, due to the $\fmax{}$ function in the computation of the $Q_1$ upper bound, the sub-gradient is non-zero for only one of the contact points, which is less efficient for training. To resolve this problem, it has been shown in \cite{Dai2018} that sum-of-squares (SOS) optimization can be used to compute a lower bound of $Q_1$. This theory can be extended to compute our generalized $Q_1$ metric. If we define $\TAN_{1,\cdots,T}(\PT_i)$ as a set of directions on the tangent plane, then the generalized $Q_1$ can be found by solving the following SOS optimization problem:
\begin{small}
\begin{align}
\label{eq:Q1L}
\argmax{}&Q_1\\
\ST&b-Q_1-L_1(\WR,b)(\WR^T\METRIC\WR-1)-\nonumber    \\
   &\sum_{ik}L_2^{ik}(\WR,b)(\SDPBASIS_{ik}^T\WR+b)\in\text{SOS}\nonumber    \\
   &L_2^{ik}(\WR,b)\in\text{SOS}\nonumber    \\
   &\SDPBASIS_{ik}\triangleq\left[\NOR(\PT_i)+\mu\TAN_k(\PT_i)\right]\nonumber   \\
   &\quad\quad\quad\EEXP(-\alpha\|\DIST(\PT_i)\|-\beta(1+\NOR(\PT_i)^T\NOR_g(\PT_i)))\nonumber,
\end{align}
\end{small}
where we have extended the definition of $\SDPBASIS_{jk}$ to account for our generalization (\prettyref{eq:Q1EXT}). \prettyref{eq:Q1L} can be reduced to a semidefinite programming (SDP) problem, and its gradients can be computed via the chain rule:
\begin{align*}
\FPP{Q_1}{\PT_i}=\FPP{Q_1}{\SDPBASIS_{ik}}\FPP{\SDPBASIS_{ik}}{\PT_i}.
\end{align*}
While the second term in the chain rule above can be computed directly via automatic differentiation, the first term $\FPPROW{Q_1}{\SDPBASIS_{ik}}$ requires a sensitivity analysis of an SDP problem, as shown in \cite{miller1997sensitivity} (see supplementary material for more details). Since SDP is a smooth approximation of a non-smooth optimization, the derivatives are generally non-zero on all the contact points. As a result, each neural network update can adjust all the fingers of the gripper to generate better grasp poses, which is more efficient than the case with an upper bound on $Q_1$. On the other hand, the cost of solving \prettyref{eq:Q1L} is also higher than that of solving \prettyref{eq:Q1U} because \prettyref{eq:Q1L} involves an SDP solve. Note that a similar analysis for quadratic programming (QP) problems has been previously exploited for training neural networks in \cite{amos2017optnet}.
\subsection{Geometry Related Loss Functions}
In this section, we show that geometric terms such as $\DIST(\PT)$ can be computed robustly from a triangle mesh. We also formulate the collision-free requirement as a novel loss term. Geometric terms arise in many places in a grasping system. To compute the $Q_1$ metric, we need to evaluate $\DIST(\PT_i)$ and $\NOR(\PT_i)$. In addition, we need to avoid penetrations between grippers and the target objects. To perform these computations, we can introduce a monotonic loss function:
\begin{align*}
\LOSS_{coll}=\EEXP(-\beta\E{min}\left[\DIST(\PT_i),0\right]^2),
\end{align*}
where $\beta$ is the weight of loss. To provide sub-gradients for all these terms, we need to plug $\FPP{\DIST(\PT_i)}{\PT_i}$ into the chain rule. In this section, we show a robust method to compute $\FPP{\DIST(\PT_i)}{\PT_i}$ for complex, watertight, triangle meshes of the target objects, which can be accelerated with the help of a bounding volume hierarchy (BVH). Note that it is easy to compute $\DIST(\PT_i)$ and its gradients from a signed distance field (SDF) \cite{baerentzen2005robust}, but we choose to use triangle meshes for two reasons. First, most existing 3D shape datasets such as \cite{calli2015benchmarking,chang2015shapenet,zhou2016thingi10k} use triangle meshes, and converting them to SDFs is time- and memory- consuming. Second, for very complex meshes, low-resolution SDFs cannot represent thin geometric features and determining an appropriate resolution of SDF is difficult.

\begin{figure}[ht]
\centering
\includegraphics[width=0.3\textwidth]{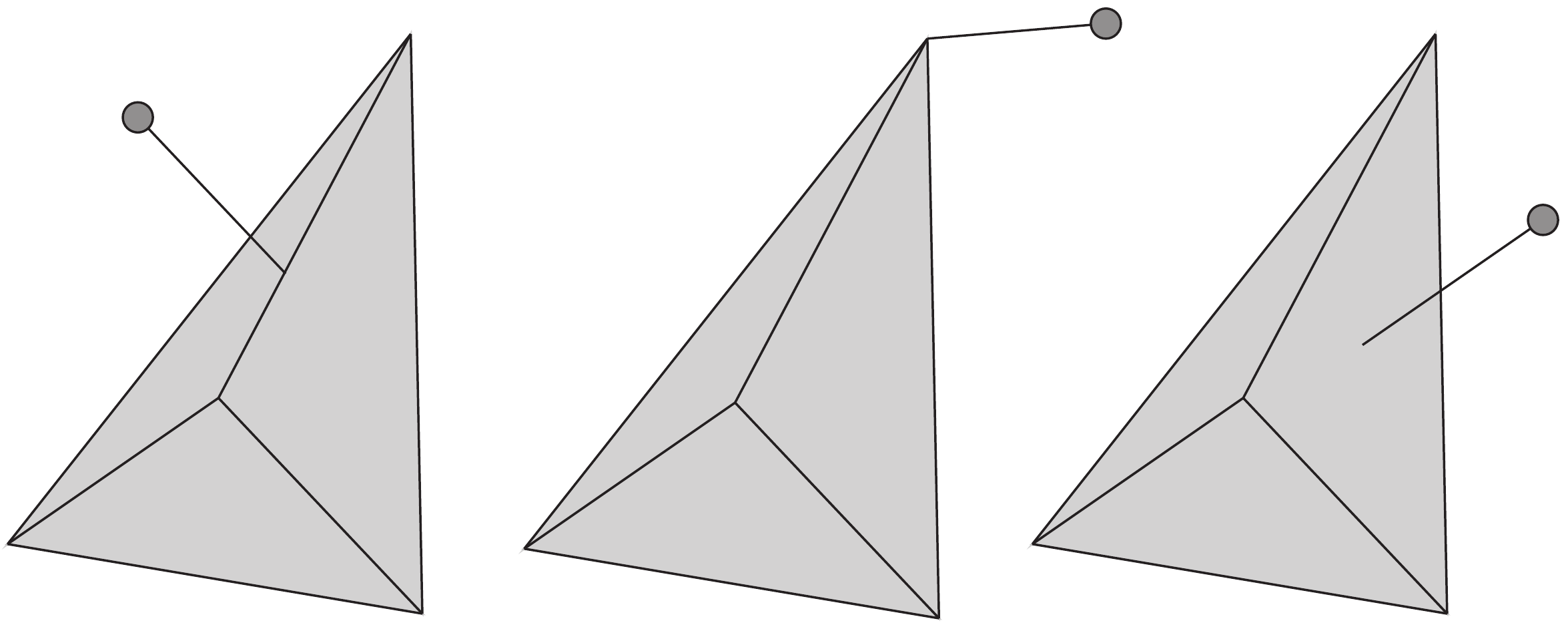}
\put(-125,30){$\EDGE$}
\put(-70,57.5){$\VERTEX$}
\put(-150,30){(a)}
\put(-100,30){(b)}
\put(-49 ,30){(c)}
\caption{\label{fig:hessian} Three cases in the computation of $\FPPROW{\DIST(\PT_i,\TRI)}{\PT_i}$. (a): The geometric feature is an edge $\EDGE$. (b): The geometric feature is a vertex $\VERTEX$. (c): The geometric feature is a triangle.}
\vspace{-20px}
\end{figure}
Let's assume that a triangle mesh $\TRI$ consists of a set of triangles $\TRI_j$. Then the distance between $\PT_i$ and $\TRI_j$ is the solution of the following QP problem:
\begin{align*}
\DIST(\PT_i,\TRI_j)=\fmin{\theta_{1,2,3}}\|\PT_i-\sum_k\theta_k\VV_{k,j}\|\quad \ST0\leq\theta_k\wedge\sum_k\theta_k\leq 1,
\end{align*}
where $\VV_{k,j}$ is the $k$th vertex of $\TRI_j$. Finally, the signed distance $\DIST(\PT_i)$ is defined as:
\begin{align}
\label{eq:SDIST}
\DIST(\PT_i)=&\DIST(\PT_i,\TRI_{j^*})
\E{sgn}(\NOR_{j^*}^T\left[\PT_i-\sum_k\theta_k\VV_{k,j^*}\right])  \\
&\ST j^*=\argmin{j}\DIST(\PT_i,\TRI_j),\nonumber
\end{align}
where $\NOR_j$ is the outward normal of $\TRI_j$ and $\E{sgn}$ is the sign function. Similarly, we can define the outward normal of $\NOR(\PT_i)$ to be:
\begin{small}
\begin{align}
\label{eq:SNOR}
\NOR(\PT_i)=-\frac{\PT_i-\sum_k\theta_k\VV_{k,j^*}}{\|\PT_i-\sum_k\theta_k\VV_{k,j^*}\|}
\E{sgn}(\NOR_{j^*}^T\left[\PT_i-\sum_k\theta_k\VV_{k,j^*}\right]).
\end{align}
\end{small}
In these formulations, the sign function and the $\argmin{j}$ operator define a disjoint convex set with well-defined sub-gradients. The gradient of $\DIST(\PT_i,\TRI)$ is:
\begin{align*}
\FPP{\DIST(\PT_i,\TRI)}{\PT_i}=\NOR(\PT_i).
\end{align*}
Also, the gradient of $\NOR(\PT_i)$ can be computed from the dirichlet features on the triangle mesh to which $\PT_i$ belongs, as illustrated in \prettyref{fig:hessian}. If the closest feature to $\PT_i$ is an edge $\EDGE$, then we have:
\begin{small}
\begin{align*}
\FPP{\NOR(\PT_i)}{\PT_i}=
\frac{\left(\frac{\EDGE\EDGE^T}{\|\EDGE\|^2}-\Id+\NOR(\PT_i)\NOR(\PT_i)^T-\frac{\EDGE^T\NOR(\PT_i)\NOR(\PT_i)\EDGE^T}{\|\EDGE\|^2}\right)}{\|\DIST(\PT_i)\|}.
\end{align*}
\end{small}
If the closest feature to $\PT_i$ is a vertex $\VERTEX$, then we have:
\begin{align*}
\FPP{\NOR(\PT_i)}{\PT_i}=\frac{\NOR(\PT_i)\NOR(\PT_i)^T-\Id}{\|\DIST(\PT_i)\|}.
\end{align*}
If the closest feature to $\PT_i$ is inside a triangle, then $\FPPROW{\NOR(\PT_i)}{\PT_i}=0$.

Finally, \prettyref{eq:SDIST} and \prettyref{eq:SNOR} involve a loop over all triangles to find the one with smallest distance, which can be accelerated by building a BVH and quickly rejecting nodes where the bounding volume is further from $\PT_i$ than the current best distance \cite{cgal:atw-aabb-19b}.

In our experiments, the technique described above is computationally efficient but prone to floating-point's truncation error. If a point is close to the triangle's plane, finite-precision floating point arithmetics have difficulty deciding whether the point lies inside the triangle mesh or not. To solve this problem, we use exact rational arithmetics implemented in \cite{Granlund:2015:GMM:2911024} to perform all the computations in this section and convert the results back to inexact, finite precision floating point numbers at the end of the computation.

\subsection{Self-Collision of the Gripper}
To prevent gripper-object collisions, we add a term $\LOSS_{self}$ to penalize any collisions between different links of the gripper. Assuming the gripper has $L$ links, we first approximate the shape of each link $i$ using a convex hull $\HULL_i$ and define $\LOSS_{self}$ as:
\begin{align*}
\LOSS_{self}=-\sum_{i=1}^L\sum_{j=1}^N\E{max}(\DIST(\PT_j,\HULL_i),0),
\end{align*}
which can be trivially computed from H-representations of $\HULL_i$ and can be accelerated using a bounding volume hierarchy. In practice, we use a small set of sample points to compute the generalized $Q_1$ metric and another large set of sample points to compute $\LOSS_{self}$ to achieve better resolution of self collisions.
\subsection{Defending Against Degenerate Cases and Local Minima}
Our generalized $Q_1$ metric is similar to the standard $Q_1$ metric in that it implies force closure. However, if an initial guess for the gripper pose has no force closure, then $Q_1=0$ and no gradient information is available. In this case, we add the following heuristic term to guide the optimization to compute a force-closed pose with a high probability:
\begin{align*}
\LOSS_{guide}=\sum_{i=1}^N\|\DIST(\PT_i,\TRI)\|^2,
\end{align*}
by ensuring that all the grasp points are as close to the object as possible. In addition, our generalized $Q_1$ has many local minima due to nonlinearity and complex geometries of objects. To defend our neural network against these sub-optimal solutions, we add a data loss to guide the training. We use Chamfer loss for our data term:
\begin{align*}
\theta^*=\argmin{\theta}\DISTGRASP(\NEURAL(D_{1,\cdots,K};\theta),\NEURAL^*),
\end{align*}
following previous works \cite{8304630,liu2019grasp}, where $\NEURAL^*$ is the ground truth grasp pose and $\DISTGRASP$ is the Chamfer distance measure in the gripper's configuration space. In other words, we precompute many ground truth grasp poses for each target object and let the neural network pick the grasp pose that leads to the minimal distance.
\subsection{Forward Kinematics}
Our neural network predicts $\NEURAL$, which consists of the global rigid transformation and the joint angles to define the pose of a gripper. Further, the gradient with respect to the grasp points $\PT_i$ is propagated backward to $\NEURAL$ via a forward kinematics layer denoted as $\E{FK}$, similar to \cite{liu2019grasp,villegas2018neural}. We make a minor modification to account for joint limits with non-vanishing gradients. If $\NEURAL$ has joint limits in range $[\E{l},\E{u}]$, then we transform $\NEURAL$ as follows:
\begin{align*}
(\E{u}-\E{l})\E{sigmoid}(\NEURAL)+\E{l},
\end{align*}
which is guaranteed to satisfy the constraints and has non-vanishing gradients compared with the $\E{min},\E{max}$ functions. 

In summary, our learning system uses the following compound loss function:
\begin{align*}
\LOSS(\TRI)\triangleq
&[\E{e}^{-Q_1}+\COEF_{coll}\LOSS_{coll}+\COEF_{self}\LOSS_{self}+\COEF_{guide}\LOSS_{guide}]\circ  \\
&\E{FK}\circ[(\E{u}-\E{l})\E{sigmoid}(\NEURAL)+\E{l}]+\COEF_{data}\DISTGRASP(\NEURAL),
\end{align*}
where $\COEF_{\bullet}$ are various weights.
\section{\label{sec:experiment}Experimental Setup}
\TE{Data Preparation:} Following \cite{8304630}, we prepare a small dataset of 500 watertight objects by combining existing grasping datasets \cite{calli2015benchmarking,singh2014bigbird,kasper2012kit,2015_ICRA_kbs}. We split the dataset into an $80\%$ (400) training set and a $20\%$ (100) test set. It is known that predicting a single grasp pose from a single target object is an ambiguous problem because many grasp poses are equally effective \cite{liu2019grasp}. Therefore, we use \cite{miller2000graspit} to precompute a set of $100$ grasp poses for each target object and then use Chamfer data loss to let the neural network pick which grasp pose is the most representable \lmr{(details can be found in \cite{liu2019grasp})}. This gives a dataset of $40$K grasps, from which our neural network will select $400$ as ground truth. For our 24-DOF gripper, collecting these data requires about 150 CPU hours of computation on a cluster using a sampling-based grasp planner \cite{miller2000graspit}. Finally, we assume that the neural network observes objects from a set of $5$ multi-view depth cameras of resolution $224\times224$. These images are obtained by \lmr{using Blender 2.79 \cite{blender} to render} the triangle mesh of each target object into the depth channel . As a result, each sample in our dataset is a $<D_{1,\cdots,5},\TRI,\NEURAL^*>$-tuple of depth images, triangle mesh, and ground truth grasp poses. After collecting our dataset, we augment it by rotating each target object and gripper for 8 times along 8 symmetric axes.

\TE{Gripper Setup:} In all our simulated and real-world experiments, we use a (6+18)-DOF Shadow Hand as our gripper, as shown in \prettyref{fig:gripper}, which is mounted onto a UR10 arm. However, during the training phase, the DOFs of the arm are not predicted by our neural network. These DOFs are computed at runtime using a conventional motion planner. We use the SrArmCommander \cite{shadowdocs} to move the UR10 arm to the target poses and use the SrHandCommander \cite{shadowdocs} to move the  Shadow Hand fingers to the target joint states. During the training phase, we manually label $N$=45 potential grasp points on the gripper and, to detect self-collisions, we use a denser sample of 15,555 potential contact points using Poisson disk sampling, as illustrated in \prettyref{fig:gripper}.
\begin{figure}[ht]
\centering
\includegraphics[width=0.3\textwidth]{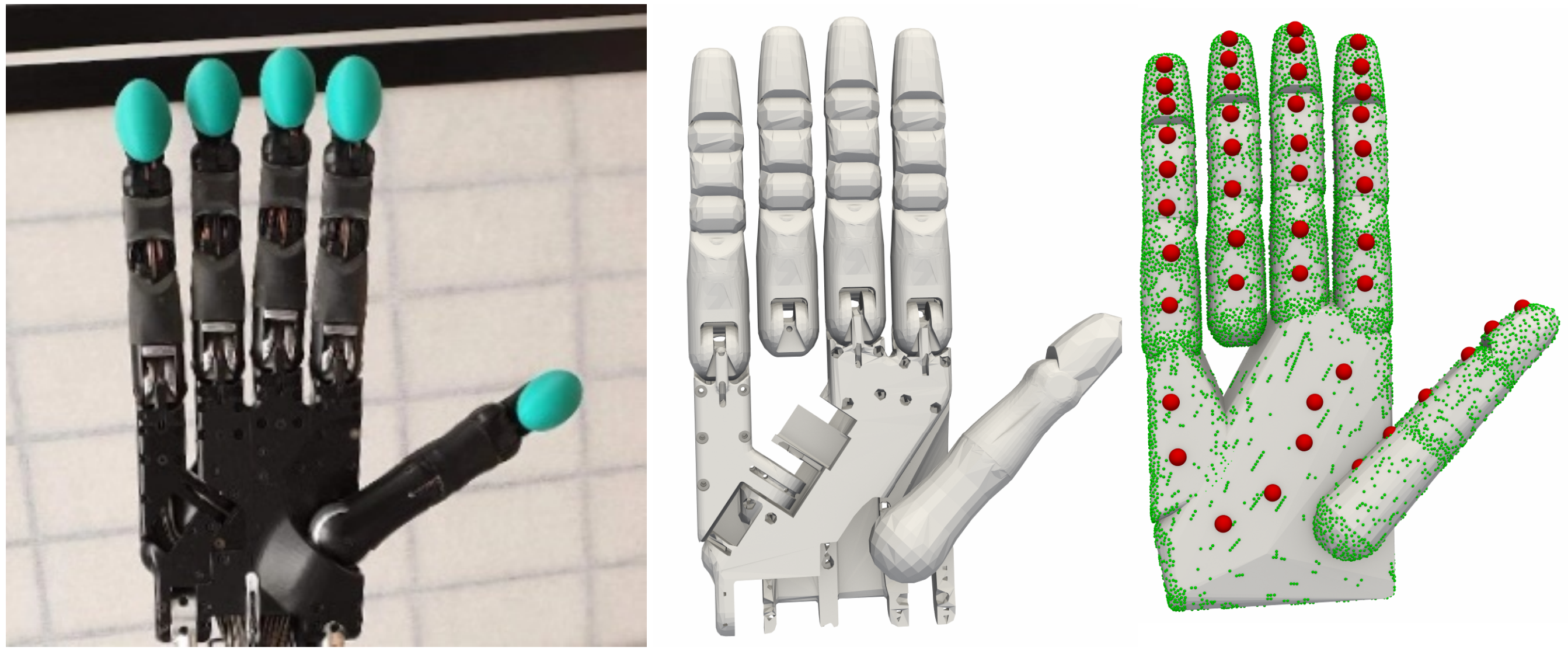}
\caption{\label{fig:gripper} Left: The real Shadow Hand. Middle: Original meshes of the Shadow Hand. Right: Convex hulls of each part of the Shadow Hand meshes, the sampled potential grasp points (red), and the sampled potential contact points (green) via Poisson disk sampling.}
\vspace{-10px}
\end{figure}

\TE{Neural Network:} We deploy a pre-trained ResNet-50 \lmr{from the TORCHVISION.MODELS offered by PyTorch \cite{paszke2017automatic}} as a feature extractor for multi-view depth images. \lmr{We then fine-tune it with depth images.} For each depth image, we duplicate it to 3 channels to meet the input requirement of ResNet-50. A shared ResNet-50 takes multi-view depth images as input and outputs 2,048 dimensional vectors. These vectors are used with max-pooling and connected with a fully-connected layer, of which the output dimension is equal to the gripper's DOF (6+18 for Shadow Hand).  Outputs of the fully-connected layer are the predicted gripper configurations.

\TE{Training configurations:} We use the parameters listed in \prettyref{table:param} for $\LOSS$ in both settings. Our neural network is trained using the ADAM algorithm \cite{kingma2014adam} with a batch size of 16. The initial learning rate is set to be 1$e$-4 and decayed by 0.9 every 20 epochs. All experiments are carried out on a desktop with 2 Intel$^{\circledR}$ Xeon Silver 4208 CPUs, 32 GB RAM, and 2 NVIDIA$^{\circledR}$ RTX 2080 GPUs.
\begin{table}[h]
\centering
\begin{tabular}{|p{1.1cm}<{\centering}|p{0.2cm}<{\centering}|p{0.2cm}<{\centering}|p{0.2cm}<{\centering}|p{0.5cm}<{\centering}|p{0.2cm}<{\centering}|p{0.4cm}<{\centering}|p{0.4cm}<{\centering}|p{0.6cm}<{\centering}|p{0.5cm}<{\centering}|}
\hline
Parameters &$\alpha$ &$\beta$ &$\mu$ &m &$D$ &$\COEF_{coll}$ &$\COEF_{self}$ &$\COEF_{guide}$ &$\COEF_{data}$\\ \hline
Value & 6.0 &8.0 &0.7 &0.001 &64 &1.0 &1.0 &0.1 &1.0 \\ \hline
\end{tabular}
\captionsetup{font={small}}
\caption{Parameter settings in our training configuration.}
\label{table:param}
\vspace{-15px}
\end{table}

\section{\label{sec:results}Experimental Results}
In this section, we evaluate the performance of different settings for high-DOF grasp planning. Our method can be used either as a standalone grasp planner or as a method to train grasp predicting neural networks.
\begin{wrapfigure}{r}{1.5cm}
\begin{center}
\includegraphics[width=1.5cm]{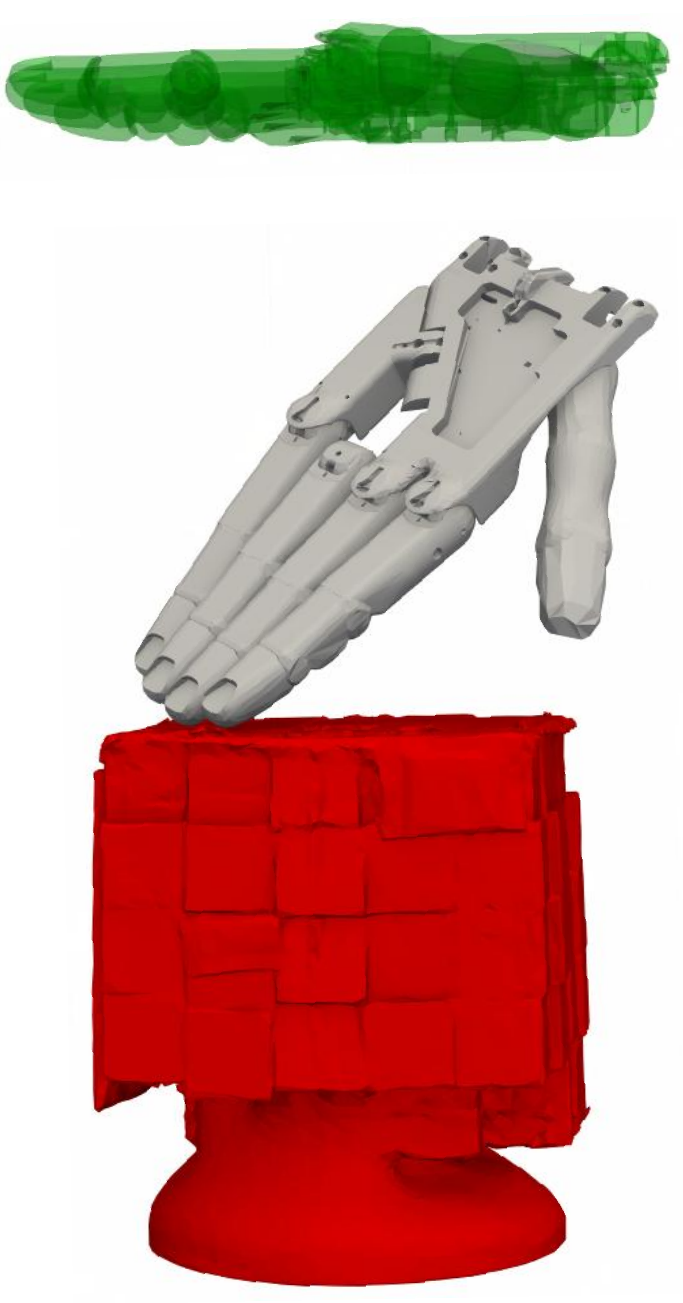}
\end{center}
\vspace{-25px}
\end{wrapfigure}
\subsection{\label{sub:only-hand}Grasp Planning without Ground Truth}
The differentiable grasp metric and collision loss allows our method to be used as a standalone, locally optimal grasp planner. To setup this experiment, we replace the neural network with a $(6+18)$-DOF optimizable vector of the gripper pose, set $\COEF_{data}=0$, and minimize $\LOSS$ with respect to $\NEURAL$. Compared to \cite{miller2000graspit}, our planner only provides \lmr{local optima}, and the computational cost is comparable. An example is illustrated in \prettyref{fig:hand_mode_more}, where we use a trivial initialization shown as the transparent green poses. After $2$ minutes of optimization, our optimizer converges to the gray poses. However, without guidance from data, our planner can fall into local minima without force closure ($Q_1=0$), as shown in the inset.

\begin{figure}[t]
\centering
\includegraphics[width=0.4\textwidth]{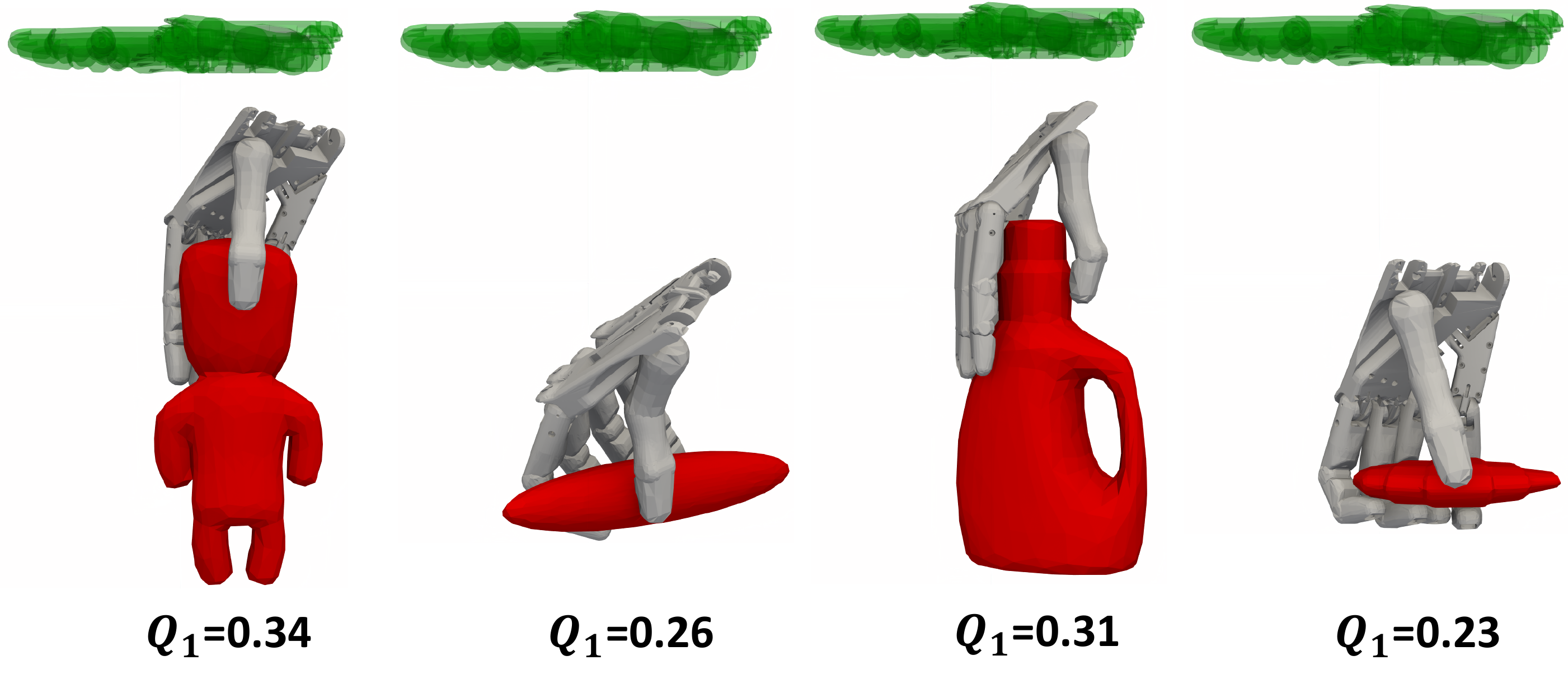}
\caption{\label{fig:hand_mode_more} We optimize the gripper pose without ground truth data. Initial pose is shown in transparent green, where all the joint angles are set to $0$ and we position the gripper directly above the object. The final poses are shown in solid gray.}
\vspace{-15px}
\end{figure}


\begin{SCfigure*}[][t]
\centering
\scalebox{0.68}{\includegraphics[width=0.97\textwidth]{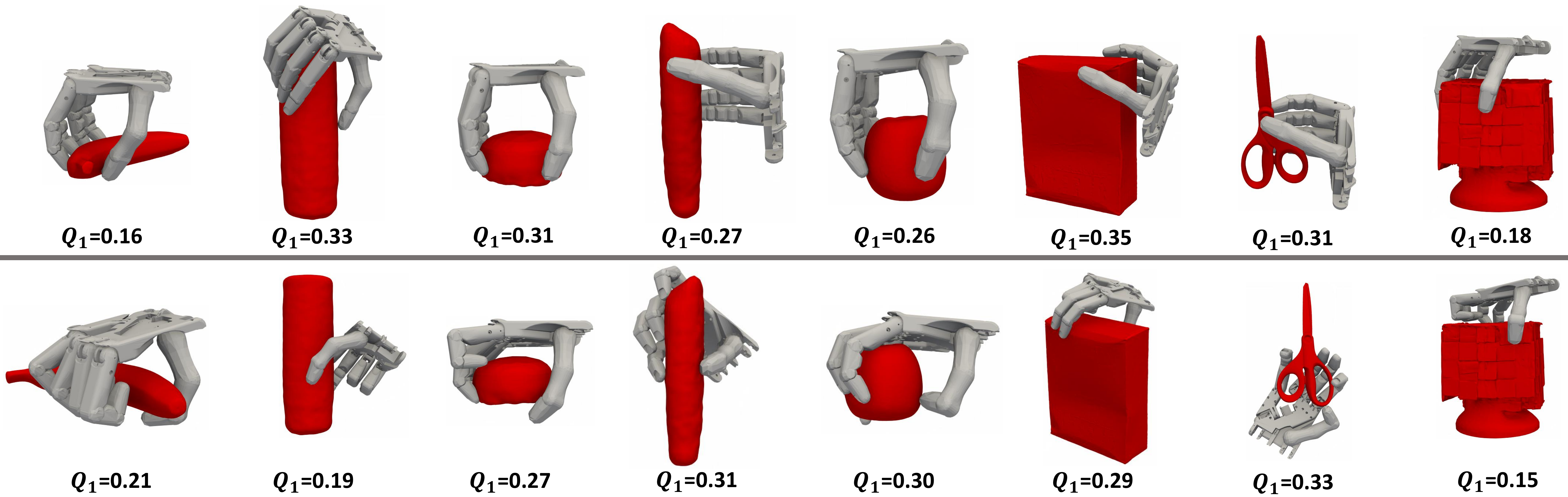}}
\caption{\small{\label{fig:NN_mode_more} Guided by the small dataset, we train our neural network by first pre-training on the dataset using Chamfer loss and then fine-tuning using our method as additional loss functions. Some predicted grasp poses for unseen objects are shown \lmr{(top row)}. These grasp poses do not require post-processing and can be realized directly on the physical platform. \lmr{For comparison, we show results of \cite{miller2000graspit} (bottom row).}}}
\end{SCfigure*}

\subsection{\label{sub:with-network}Learning Grasp Poses with Ground Truth}
In our second benchmark, we use our method to guide the training of a grasp-pose-predicting neural network. The training is performed in two phases. First, we adopt a pre-training by setting $\LOSS=\LOSS_{data}$, i.e. excluding our differentiable loss. This step brings the neural network close to nearly optimal values and we run 35 epochs of learning at the first stage. Second, we fine-tune the network by adding our differentiable loss and use weights as summarized in \prettyref{table:param}. We run 71 epochs of learning at the second stage. The pre-training takes 4 hours and the fine-tuning takes 36 hours. On average, each forward-backward propagation with our additional loss function takes 0.85s and the one without our loss function takes 0.61s, which shows that our additional loss functions only impose a marginal cost to gradient computation. \lmr{However, to ensure fine-grained convergence to a good local minimum, we use a small learning rate and more epochs for the second stage, which runs $9\times$ slower than the first stage.} After training, \lmr{we test our neural network on the set of 100 held-out objects, on which the mean $Q_1$ metric is 0.226 and the variance of the $Q_1$ metric is 0.0045.} A set of predicted grasp poses on the test set is shown in \prettyref{fig:post-process}, from which we observe drastically improved grasp quality when guided by the data term. \lmr{On a physical platform, however, there might be environmental constraints making our predicted grasps infeasible. In this case, we can randomly perturb object poses to create virtual depth images and predict a set of varied grasps from them, as shown in \prettyref{fig:different_pose}, from which we can pick one feasible grasp.}


\begin{SCfigure}[][h]
\centering
\scalebox{0.33}{\includegraphics[width=0.95\textwidth]{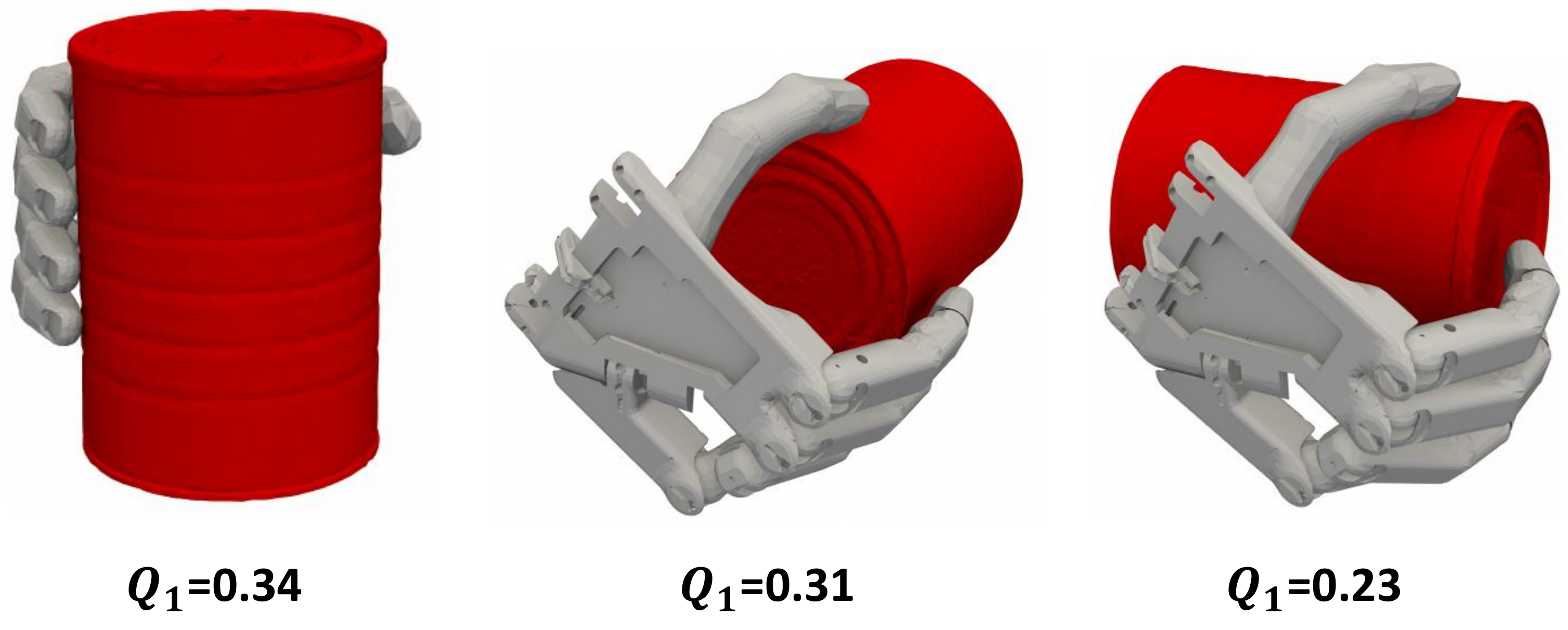}}
\caption{\small{\label{fig:different_pose} \lmr{By changing the orientation of the object, we generate a set of varied grasps, from which we pick feasible grasps.}}}
\end{SCfigure}

\subsection{Comparison}
We have compared the (standard) $Q_1$ metric \cite{219918} of our method and \lmr{a sampling-based grasp planner} \cite{miller2000graspit} in \prettyref{fig:NN_mode_more}. The results show that the qualities of our grasp poses are on par with those of \cite{miller2000graspit}. We have also compared our approach with prior work \cite{liu2019grasp}, which also trains a grasp-pose predicting neural network on a small dataset of a  size similar to ours. However, the algorithm in \cite{liu2019grasp} requires a post-processing step to resolve penetrations and collisions. Instead, the grasp poses predicted using our method can be directly deployed onto a physical hardware without post-processing. As illustrated in \prettyref{fig:post-process} and \prettyref{table:comp_liu}, our method can significantly improve the quality of grasp poses.

\begin{SCfigure}[][h]
\centering
\scalebox{0.76}{\includegraphics[width=0.4\textwidth]{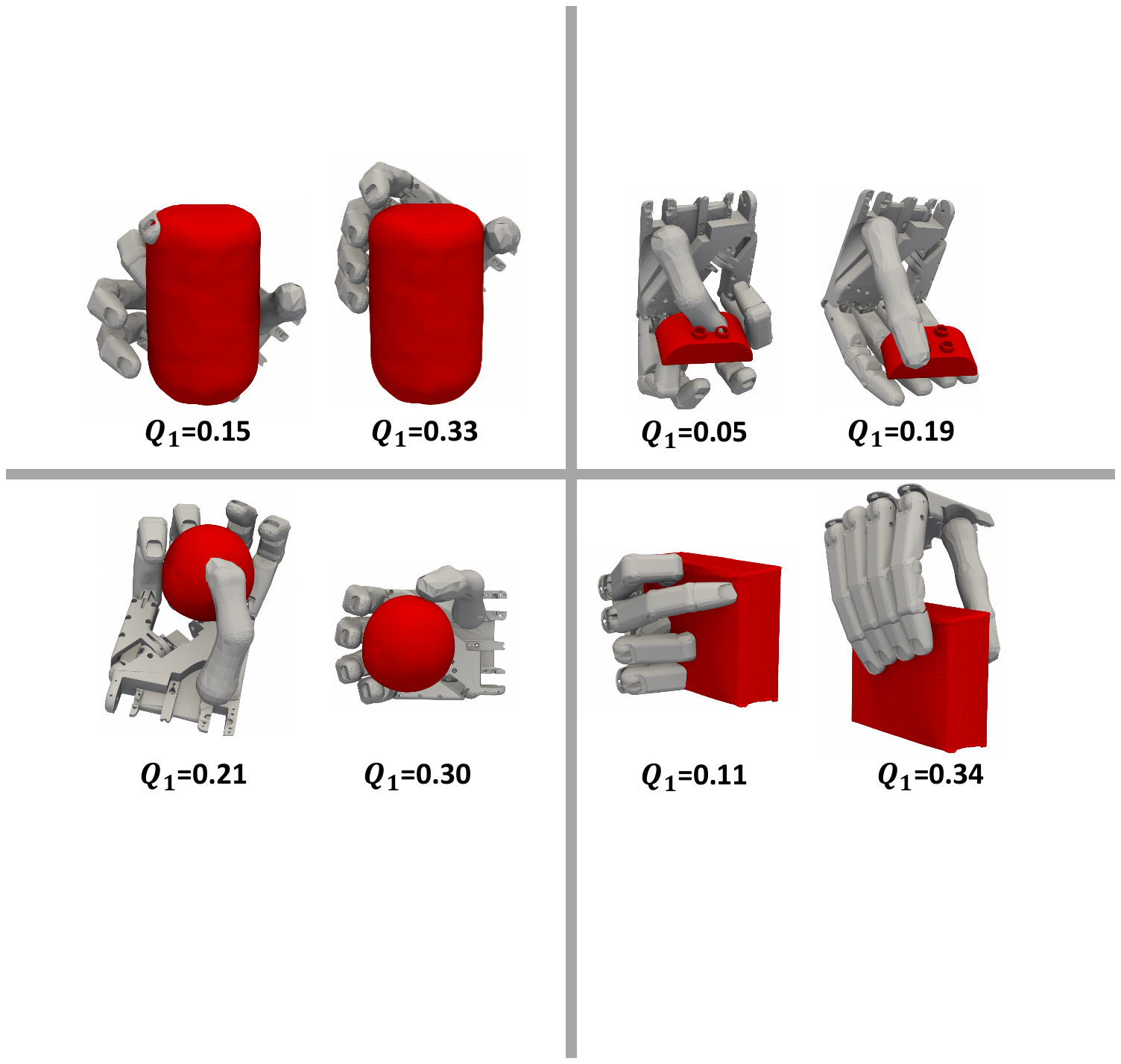}}
\caption{\label{fig:post-process} We compare our method (Right) and prior work \cite{liu2019grasp} (Left). Our method can drastically improve the quality of grasps and reduce penetrations or collisions.}
\vspace{-15px}
\end{SCfigure}

\subsection{Grasping with the Arm on Physical Hardware}
As our final evaluation, we deploy our learned neural network onto our physical platform. Our method does not require RGB input and only uses the depth channel. Therefore, we do not perform any sim-to-real transfer. Our neural network only predicts the gripper pose and does not predict the configuration for the UR10 arm to achieve the predicted position and orientation. These configurations of the arm are computed using a motion planner at runtime. We choose 50 YCB objects from our 100 test objects. \lmr{All YCB objects are unseen and excluded from the training set. We use the depth images from five ASUS Xtion PRO LIVE cameras with 640x480 resolution as our network input.} Our depth cameras are calibrated beforehand to make the camera pose exactly the same as in training. \lmr{We then crop the real depth images and remove the background.} \lmr{Moreover, we make objects' poses exactly the same as the poses used for rendering depth images.}
\begin{table}[h]
\centering
\vspace{-5px}
\scalebox{0.9}{
\begin{tabular}{|m{8mm}<{\centering}|m{7.5mm}<{\centering}|m{12mm}<{\centering}|m{8mm}<{\centering}|m{7.5mm}<{\centering}|m{7.5mm}<{\centering}|m{15mm}<{\centering}|}
\hline
Method& $Q_1$ Metric & Penetration & Success Plan & Success Grasp & \lmr{Success Rate} & \lmr{Overall Success Rate} \\
\hline
Ours &0.23 &3.5mm &38 & 33 & \lmr{86.8\%} & \lmr{66.0\%}\\
\hline
\cite{liu2019grasp} &0.11 &14.2mm &35 & 27 & \lmr{77.1\%} & \lmr{54.0\%} \\
\hline
\end{tabular}}
\captionsetup{font={small}}
\caption{For the $50$ YCB objects in the testing set, we compare the predicted quality of grasp poses in terms of the $Q_1$ metric, penetration depth, and rate of success for the planner and physical hardware.}
\vspace{-15px}
\label{table:comp_liu}
\end{table}

To profile the rate of success on the 50 YCB objects, we use two metrics summarized in \prettyref{table:comp_liu}. First, we record how many times the motion planner can successfully move the gripper to the predicted position (Success-Plan). This metric measures the ability of our method to avoid penetrations and collisions with the desk on which the objects are placed since a pose with penetrations or desk collisions cannot be achieved by a motion planner. Second, we record how many times the grasp planner can successfully lift the object (Success-Grasp). \lmr{We define our Success-Rate as the percentage of Success-Grasp out of Success-Plan, and we define Overall-Success-Rate as the percentage of Success-Grasp out of the 50 trials.} This metric measures the ability of our method to improve the grasp quality. Our method outperforms \cite{liu2019grasp} in terms of both metrics. We observe an $8\%$ improvement in terms of Success Plan and a $22\%$ improvement in terms of Success Grasp. \lmr{We claim that we have succeeded when the object has no relative motion against the hand for a sufficiently long period.} Our neural network failed on $5$ objects due to slippage. \lmr{These 5 objects are: wood\_block, power\_drill, extra\_large\_clamp, hammer, and potted\_meat\_can.}

\section{\label{sec:conclusion}Conclusion and limitations}
We present a differentiable grasp planner that enables a neural network to be trained with a small dataset and a simplified architecture. Our differentiable loss accounts for various requirements for a good grasp, including high grasp metric values and collision-free gripper poses. We use a generalized definition to allow inexact contact and we show that the sub-gradients of each loss term are well-defined and can be efficiently computed from target object shapes represented using watertight triangle meshes. We show that our method can be used both as a standalone grasp planner and as a neural network training algorithm. Finally, we show that the trained neural network performs robustly on unseen objects and hardware platforms.

Our current implementation suffers from several limitations. First, our method requires the target objects to be watertight and to have a non-zero volume. Although we do not require a signed distance field transformation, our method still computes a signed value of distance, which is impossible when the target object is a thin-shell. A limitation related to this problem is that our method suffers from tunneling. In other words, when the target object is very thin, a stochastic update of our neural network might result in the hand going from one side to the other side of the object, leading to missed solutions. In the future, this problem can be resolved using continuous collision detection \cite{brochu2012efficient}. Second, our experimental setup and neural network architecture prevents the neural network from predicting multiple grasp poses for a single object. If there are other constraints in the workspace preventing a grasp pose from being achieved, then our method will lead to failure. However, this problem can be resolved by using adversarial training similar to \cite{fang2018mtda,Mousavian_2019_ICCV}, where a distribution of grasp poses is learned. We emphasize that more sophisticated learning algorithms are orthogonal to our approach and can be combined with it. Finally, by using the exact $Q_1$ metric as our loss function, we can only generate precision grasps, meaning more robust power grasp or caging grasp generation is left as future work.
\section{Acknowledgments}
{We thank the anonymous reviewers for their valuable comments. This work was supported in part by the National Key Research and Development Program of China (No. 2018AAA0102200), NSFC (61572507, 61532003, 61622212), ARO grant W911NF-18-1-0313, and Intel. Min Liu is supported by the China Scholarship Council.}

\clearpage
\onecolumn
\section{\label{sec:appendix}Appendix}
In this document, we provide some details on computing the $Q_1$ lower bound and its derivatives. First, we derive a slightly different formulation of the $Q_1$ lower bound using quadratic frictional cones. As compared with the linearized frictional cones used in \cite{Dai2018}, using quadratic frictional cones is more efficient in terms of reducing the problem size of semidefinite programming.

\textbf{1  Q1 Lower Bound Using Quadratic Frictional Cone}

We re-derive the lower bound of $Q_1$ using SOS optimization as done in \cite{Dai2018}, but using quadratic frictional cones. For a set of points $\PT_{1,\cdots,N}$, with normals $\NOR(\PT_i)$ and two tangents being $\TAN_{1,2}$, then the cones $\mathcal{K}_{\mathcal{B},\mathcal{W}}$ are defined as:
\begin{align*}
\mathcal{K}_\mathcal{B}&\triangleq\{\TWOC{\WR}{t}|\WR^T\METRIC\WR\leq r^2t^2,t\geq0\} \\
\mathcal{K}_\mathcal{W}&\triangleq\{\TWOC{\SDPBASIS}{\sum_i\lambda_i}|
\sum_i\SDPBASIS_i^\perp\lambda_i+\SDPBASIS_i^{\parallel1}\alpha_i+
\SDPBASIS_i^{\parallel2}\beta_i,\mu\lambda_i\geq\sqrt{\alpha_i^2+\beta_i^2}\},
\end{align*}
where we have:
\begin{align*}
\SDPBASIS_i^\perp       \triangleq\TWOC{\NOR  (\PT_i)}{\PT_i\times\NOR  (\PT_i)}\quad
\SDPBASIS_i^{\parallel1}\triangleq\TWOC{\TAN_1(\PT_i)}{\PT_i\times\TAN_1(\PT_i)}\quad
\SDPBASIS_i^{\parallel2}\triangleq\TWOC{\TAN_2(\PT_i)}{\PT_i\times\TAN_2(\PT_i)}.
\end{align*}
It is easy to find that the dual cones of $\mathcal{K}_{\mathcal{B},\mathcal{W}}$ are defined as:
\begin{align*}
\mathcal{K}_\mathcal{B}^*&\triangleq\{\TWOC{\E{a}}{b}|b^2\geq r^2\E{a}^T\METRIC^{-1}\E{a},b\geq0\} \\
\mathcal{K}_\mathcal{W}^*&\triangleq\{\TWOC{\E{a}}{b}|({\SDPBASIS_i^\perp}^T\E{a}+b)/\mu\geq
\sqrt{({\SDPBASIS_i^{\parallel1}}^T\E{a})^2+({\SDPBASIS_i^{\parallel2}}^T\E{a})^2}\}.
\end{align*}
The induced SOS problem is:
\begin{align}
\label{eq:SOS}
\begin{cases}
{\SDPBASIS_i^\perp}^T\E{a}+b\geq0   \\
({\SDPBASIS_i^\perp}^T\E{a}+b)^2/\mu^2\geq
({\SDPBASIS_i^{\parallel1}}^T\E{a})^2+({\SDPBASIS_i^{\parallel2}}^T\E{a})^2 \\
\E{a}^T\METRIC^{-1}\E{a}=1
\end{cases} \Longrightarrow b>r.
\end{align}
Note that \prettyref{eq:SOS} will induce an SDP problem with exactly the same order (of polynomials) as the original SDP problem induced in \cite{Dai2018}, but with fewer cones and also smaller linear system when performing sensitivity analysis. Finally, we briefly prove the correctness of $\mathcal{K}_\mathcal{W}^*$. 
\begin{lemma}
The dual cone of $\mathcal{K}_\mathcal{W}$ is $\mathcal{K}_\mathcal{W}^*$.
\end{lemma}
\textbf{Proof:} If $\TWOR{\E{a}}{b}\in\mathcal{K}_\mathcal{W}^*$, then for any $\TWOR{\SDPBASIS}{\sum_i\lambda_i}\in\mathcal{K}_\mathcal{W}$, we have:
\begin{align*}
 \TWO{\SDPBASIS^T}{\sum_i\lambda_i}\TWOC{\E{a}}{b}
=&\sum_i
  {\SDPBASIS_i^\perp}^T\E{a}\lambda_i+
  {\SDPBASIS_i^{\parallel1}}^T\E{a}\alpha_i+
  {\SDPBASIS_i^{\parallel2}}^T\E{a}\beta_i+\lambda_ib  \\
\geq&\sum_i
  \sqrt{({\SDPBASIS_i^{\parallel1}}^T\E{a})^2+({\SDPBASIS_i^{\parallel2}}^T\E{a})^2}\lambda_i\mu+
  {\SDPBASIS_i^{\parallel1}}^T\E{a}\alpha_i+{\SDPBASIS_i^{\parallel2}}^T\E{a}\beta_i    \\
\geq&\sum_i
  \sqrt{({\SDPBASIS_i^{\parallel1}}^T\E{a})^2+({\SDPBASIS_i^{\parallel2}}^T\E{a})^2}
  \sqrt{\alpha_i^2+\beta_i^2}+
  {\SDPBASIS_i^{\parallel1}}^T\E{a}\alpha_i+{\SDPBASIS_i^{\parallel2}}^T\E{a}\beta_i\geq0.
\end{align*}
For the other direction, if there is an $i$ such that $({\SDPBASIS_i^\perp}^T\E{a}+b)/\mu<
\sqrt{({\SDPBASIS_i^{\parallel1}}^T\E{a})^2+({\SDPBASIS_i^{\parallel2}}^T\E{a})^2}$, then we can pick a point in $\mathcal{K}_\mathcal{W}$ as follows:
\begin{align*}
\TWOC{\SDPBASIS}{1}=\TWOC{\SDPBASIS_i^\perp-\mu
\frac{{\SDPBASIS_i^{\parallel1}}^T\E{a}\SDPBASIS_i^{\parallel1}+
      {\SDPBASIS_i^{\parallel2}}^T\E{a}\SDPBASIS_i^{\parallel2}}
{\sqrt{({\SDPBASIS_i^{\parallel1}}^T\E{a})^2+({\SDPBASIS_i^{\parallel2}}^T\E{a})^2}}}
{1}\in\mathcal{K}_\mathcal{W},
\end{align*}
such that:
\begin{align*}
\SDPBASIS^T\E{a}+b={\SDPBASIS_i^\perp}^T\E{a}-\mu
{\sqrt{({\SDPBASIS_i^{\parallel1}}^T\E{a})^2+({\SDPBASIS_i^{\parallel2}}^T\E{a})^2}}+b<0.
\end{align*}

\textbf{2 SDP Sensitivity Analysis for Lower Bound of $Q_1$}

In this section, we present an efficient way to perform sensitivity analysis for SOS problems. We use the same notations as those in \cite{miller1997sensitivity}. A standard SDP takes the form:
\begin{align*}
&\argmin{\XX}\CC^T\XX\quad \ST{}\FFF^j\in\text{PSD}    \\
&\FFF^j\triangleq\FFF_0^j+\sum_i\FFF_i^j\XX_i,
\end{align*}
where there are $j=1,\cdots,1+KN$ PSD cones in our problem ($K$ equals the number of tangent directions if linearized frictional cones are used and $K=2$ if quadratic frictional cones are used). The dual variable to the $j$th cone is $\ZZZ^j$ and we have $\FFF^j\ZZZ^j=0$. We also define the coefficient matrix:
\begin{align*}
\FFFF=\THREE{\FPP{\SVEC{\FFF^1}}{\XX}^T}{\hdots}{\FPP{\SVEC{\FFF^{1+KN}}}{\XX}^T}^T.
\end{align*}
In an SOS problem, the first PSD cone and the $KN$ other cones are of two different types. The first PSD cone $\FFF^1$ specifies the conditional polynomial positivity condition. The other $KN$ cones $\FFF^{KN}$ specify the positivity of Lagrangian multipliers. We also observe that some variables $\XX^1$ only affect the first PSD cone and other variables $\XX^{KN}$ affect the other $KN$ PSD cones. Therefore, we can write the matrix $\FFFF$ in a $2\times2$ block form as follows:
\begin{align*}
\FFFF=\MTT{\FPP{\SVEC{\FFF^1}}{\XX^1}}{\FPP{\SVEC{\FFF^1}}{\XX^{KN}}}{}{\Id},
\end{align*}
where it is trivial to verify that we can choose variables to make the bottom right block of $\FFFF$ an identity matrix. When SDP is solved using primal-dual interior point method, the set of primal and dual solutions are computed simultaneously, with the dual variables defined as:
\begin{align*}
\ZZZZ=\TWOR{\SVEC{\ZZZ^1}}{\SVEC{\ZZZ^{KN}}},
\end{align*}
where we apply the same decomposition of cones for $\ZZZ$. Next, we apply the optimalty condition of SDP:
\begin{align*}
G=\THREEC
{\FFFF^T\ZZZZ-\CC}
{(\ZZZ^1\otimes\Id)\SVEC{\FFF^1}}
{(\ZZZ^{KN}\otimes\Id)\SVEC{\FFF^{KN}}}=0,
\end{align*}
where $\otimes$ is the symmetric kronecker product operator. Apply sensitivity analysis with respect to an arbitrary parameter $\epsilon$, we have:
\begin{small}
\begin{align*}
&\left(\setlength{\arraycolsep}{3pt}\def\arraystretch{2}
\begin{array}{c|c|c|c}
 &  & \FPP{\SVEC{\FFF^1}}{\XX^1}^T &    \\
\hline
 &  & \FPP{\SVEC{\FFF^1}}{\XX^{KN}}^T & \Id \\
\hline
(\ZZZ^1\otimes\Id)\FPP{\SVEC{\FFF^1}}{\XX^1} & 
(\ZZZ^1\otimes\Id)\FPP{\SVEC{\FFF^1}}{\XX^{KN}} & 
\FFF^1\otimes\Id &  \\
\hline
 & \ZZZ^{KN}\otimes\Id &  & \FFF^{KN}\otimes\Id \\
\end{array}\right)
\FOURC
{\FPP{\XX^1}{\epsilon}}
{\FPP{\XX^{KN}}{\epsilon}}
{\FPP{\SVEC{\ZZZ^1}}{\epsilon}}
{\FPP{\SVEC{\ZZZ^{KN}}}{\epsilon}}+\FOURC{0}{b_1}{b_2}{0}=0,
\end{align*}
\end{small}
where:
\begin{small}
\begin{align*}
&b_1\triangleq\FPPTT{\SVEC{\FFF^1}}{\XX^{KN}}{\epsilon}^T\SVEC{\ZZZ^1}   \quad
b_2\triangleq(\ZZZ^1\otimes\Id)\FPPTT{\SVEC{\FFF^1}}{\XX^{KN}}{\epsilon}\XX^{KN}.
\end{align*}
\end{small}
In the following derivation, we assume that $\FFF^{KN}$ and $\ZZZ^{KN}$ have strict complementarity. Note that if strict complementarity is not satisfied, then the SDP problem is not differentiable. Prior work \cite{miller1997sensitivity} showed that $\FFF^{1,KN}$ and $\ZZZ^{1,KN}$ have simultaneous diagonalization, and so does $\FFF^{1,KN}\otimes\Id$ and $\ZZZ^{1,KN}\otimes\Id$:
\begin{small}
\begin{align*}
\ZZZ^1\otimes\Id=\THREEC{{V_1^1}^T}{{V_2^1}^T}{{V_3^1}^T}^T
\MDD{\Sigma_{\ZZZ1}^1}{\Sigma_{\ZZZ2}^1}{0}
\THREEC{{V_1^1}^T}{{V_2^1}^T}{{V_3^1}^T}
&\quad\FFF^1\otimes\Id=\THREEC{{V_1^1}^T}{{V_2^1}^T}{{V_3^1}^T}^T
\MDD{0}{\Sigma_{\FFF1}^1}{\Sigma_{\FFF2}^1}
\THREEC{{V_1^1}^T}{{V_2^1}^T}{{V_3^1}^T}   \\
\ZZZ^{KN}\otimes\Id=\THREEC{{V_1^{KN}}^T}{{V_2^{KN}}^T}{{V_3^{KN}}^T} ^T
\MDD{\Sigma_{\ZZZ1}^{KN}}{\Sigma_{\ZZZ2}^{KN}}{0}
\THREEC{{V_1^{KN}}^T}{{V_2^{KN}}^T}{{V_3^{KN}}^T} 
&\quad\FFF^{KN}\otimes\Id=\THREEC{{V_1^{KN}}^T}{{V_2^{KN}}^T}{{V_3^{KN}}^T}^T
\MDD{0}{\Sigma_{\FFF1}^{KN}}{\Sigma_{\FFF2}^{KN}}
\THREEC{{V_1^{KN}}^T}{{V_2^{KN}}^T}{{V_3^{KN}}^T}.
\end{align*}
\end{small}
By plugging these identities info the sensitivity equation, we get:
\begin{tiny}
\begin{align*}
&\left(\setlength{\arraycolsep}{3pt}\def\arraystretch{2}
\begin{array}{c|c|c|c|c|c|c|c}
 &  & 
\FPP{\SVEC{\FFF^1}}{\XX^1}^TV_1^1 & 
\FPP{\SVEC{\FFF^1}}{\XX^1}^TV_2^1 & 
\FPP{\SVEC{\FFF^1}}{\XX^1}^TV_3^1 &
 &  &   \\
\hline
 &  & 
\FPP{\SVEC{\FFF^1}}{\XX^{KN}}^TV_1^1 & 
\FPP{\SVEC{\FFF^1}}{\XX^{KN}}^TV_2^1 & 
\FPP{\SVEC{\FFF^1}}{\XX^{KN}}^TV_3^1 &
V_1^{KN} & V_2^{KN} & V_3^{KN} \\
\hline
\Sigma_{\ZZZ1}^1{V_1^1}^T\FPP{\SVEC{\FFF^1}}{\XX^1} & \Sigma_{\ZZZ1}^1{V_1^1}^T\FPP{\SVEC{\FFF^1}}{\XX^{KN}} & 
 &  &  &  &  &  \\
\hline
\Sigma_{\ZZZ2}^1{V_2^1}^T\FPP{\SVEC{\FFF^1}}{\XX^1} & \Sigma_{\ZZZ2}^1{V_2^1}^T\FPP{\SVEC{\FFF^1}}{\XX^{KN}} & 
 & \Sigma_{\FFF1}^1 &  &  &  &  \\
\hline
 &  &  &  & \Sigma_{\FFF2}^1 &  &  &    \\
\hline
 & \Sigma_{\ZZZ1}^{KN}{V_1^{KN}}^T &  &  &  &  &  &   \\
\hline
 & \Sigma_{\ZZZ2}^{KN}{V_2^{KN}}^T &  &  &  &  & \Sigma_{\FFF1}^{KN} &    \\
\hline
 &  &  &  &  &  &  & \Sigma_{\FFF2}^{KN}    \\
\end{array}\right)  \\
&\EIGHTC
{\FPP{\XX^1}{\epsilon}}
{\FPP{\XX^{KN}}{\epsilon}}
{{V_1^1}^T\FPP{\SVEC{\ZZZ^1}}{\epsilon}}
{{V_2^1}^T\FPP{\SVEC{\ZZZ^1}}{\epsilon}}
{{V_3^1}^T\FPP{\SVEC{\ZZZ^1}}{\epsilon}}
{{V_1^{KN}}^T\FPP{\SVEC{\ZZZ^{KN}}}{\epsilon}}
{{V_2^{KN}}^T\FPP{\SVEC{\ZZZ^{KN}}}{\epsilon}}
{{V_3^{KN}}^T\FPP{\SVEC{\ZZZ^{KN}}}{\epsilon}}+
\EIGHTC
{0}
{b_1}
{{V_1^1}^Tb_2}
{{V_2^1}^Tb_2}
{0}
{0}
{0}
{0}=0,
\end{align*}
\end{tiny}
where there are 8 equations. For $4,5,7,8$th rows, we have:
\begin{small}
\begin{equation}
\begin{aligned}
\label{eq:reduction}
{V_2^1}^T\FPP{\SVEC{\ZZZ^1}}{\epsilon}&=
-{\Sigma_{\FFF1}^1}^{-1}\Sigma_{\ZZZ2}^1{V_2^1}^T\FPP{\SVEC{\FFF^1}}{\XX^1}\FPP{\XX^1}{\epsilon}    \\
&-{\Sigma_{\FFF1}^1}^{-1}\Sigma_{\ZZZ2}^1{V_2^1}^T\FPP{\SVEC{\FFF^1}}{\XX^{KN}}\FPP{\XX^{KN}}{\epsilon} \\
&-{\Sigma_{\FFF1}^1}^{-1}{V_2^1}^Tb_2    \\
{V_3^1}^T\FPP{\SVEC{\ZZZ^1}}{\epsilon}&=0   \\
{V_2^{KN}}^T\FPP{\SVEC{\ZZZ^{KN}}}{\epsilon}&=
-{\Sigma_{\FFF1}^{KN}}^{-1}\Sigma_{\ZZZ2}^{KN}{V_2^{KN}}^T\FPP{\XX^{KN}}{\epsilon}   \\
{V_3^{KN}}^T\FPP{\SVEC{\ZZZ^{KN}}}{\epsilon}&=0.
\end{aligned}
\end{equation}
\end{small}
By plugging \prettyref{eq:reduction} into the $1$st row, we have:
\begin{small}
\begin{align*}
&\FPP{\SVEC{\FFF^1}}{\XX^1}^TV_1^1({V_1^1}^T\FPP{\SVEC{\ZZZ^1}}{\epsilon})-  \\
&\FPP{\SVEC{\FFF^1}}{\XX^1}^TV_2^1
({\Sigma_{\FFF1}^1}^{-1}\Sigma_{\ZZZ2}^1{V_2^1}^T\FPP{\SVEC{\FFF^1}}{\XX^1}\FPP{\XX^1}{\epsilon}+
{\Sigma_{\FFF1}^1}^{-1}\Sigma_{\ZZZ2}^1{V_2^1}^T\FPP{\SVEC{\FFF^1}}{\XX^{KN}}\FPP{\XX^{KN}}{\epsilon}+
{\Sigma_{\FFF1}^1}^{-1}{V_2^1}^Tb_2)=0.
\end{align*}
\end{small}
By plugging \prettyref{eq:reduction} into the $2$nd row, we have:
\begin{small}
\begin{align*}
&\FPP{\SVEC{\FFF^1}}{\XX^{KN}}^TV_1^1({V_1^1}^T\FPP{\SVEC{\ZZZ^1}}{\epsilon})-  \\
&\FPP{\SVEC{\FFF^1}}{\XX^{KN}}^TV_2^1
({\Sigma_{\FFF1}^1}^{-1}\Sigma_{\ZZZ2}^1{V_2^1}^T\FPP{\SVEC{\FFF^1}}{\XX^1}\FPP{\XX^1}{\epsilon}+
{\Sigma_{\FFF1}^1}^{-1}\Sigma_{\ZZZ2}^1{V_2^1}^T\FPP{\SVEC{\FFF^1}}{\XX^{KN}}\FPP{\XX^{KN}}{\epsilon}+
{\Sigma_{\FFF1}^1}^{-1}{V_2^1}^Tb_2)+  \\
&V_1^{KN}{V_1^{KN}}^T\FPP{\SVEC{\ZZZ^1}}{\epsilon}-
V_2^{KN}{\Sigma_{\FFF1}^{KN}}^{-1}\Sigma_{\ZZZ2}^{KN}{V_2^{KN}}^T\FPP{\XX^{KN}}{\epsilon}+b_1=0.
\end{align*}
\end{small}
The $3,6$th rows will remain intact and $6$th row can be eliminated. Finally, our reduced sensitivity equation is:
\begin{tiny}
\begin{align*}
&\left(\setlength{\arraycolsep}{3pt}\def\arraystretch{2}
\begin{array}{c|c|c|c}
-\FPP{\SVEC{\FFF^1}}{\XX^1}^TV_2^1
{\Sigma_{\FFF1}^1}^{-1}\Sigma_{\ZZZ2}^1{V_2^1}^T\FPP{\SVEC{\FFF^1}}{\XX^1} & 
-\FPP{\SVEC{\FFF^1}}{\XX^1}^TV_2^1
{\Sigma_{\FFF1}^1}^{-1}\Sigma_{\ZZZ2}^1{V_2^1}^T\FPP{\SVEC{\FFF^1}}{\XX^{KN}}& 
\FPP{\SVEC{\FFF^1}}{\XX^1}^TV_1^1 & \\
\hline
 & -\FPP{\SVEC{\FFF^1}}{\XX^{KN}}^TV_2^1
{\Sigma_{\FFF1}^1}^{-1}\Sigma_{\ZZZ2}^1{V_2^1}^T\FPP{\SVEC{\FFF^1}}{\XX^{KN}} & &   \\
\smash{\raisebox{1\normalbaselineskip}{$-\FPP{\SVEC{\FFF^1}}{\XX^{KN}}^TV_2^1
{\Sigma_{\FFF1}^1}^{-1}\Sigma_{\ZZZ2}^1{V_2^1}^T\FPP{\SVEC{\FFF^1}}{\XX^1}$}} &
-V_2^{KN}{\Sigma_{\FFF1}^{KN}}^{-1}\Sigma_{\ZZZ2}^{KN}{V_2^{KN}}^T &
\smash{\raisebox{1\normalbaselineskip}{$\FPP{\SVEC{\FFF^1}}{\XX^{KN}}^TV_1^1$}} &
\smash{\raisebox{1\normalbaselineskip}{$V_1^{KN}$}} \\
\hline
\Sigma_{\ZZZ1}^1{V_1^1}^T\FPP{\SVEC{\FFF^1}}{\XX^1} & \Sigma_{\ZZZ1}^1{V_1^1}^T\FPP{\SVEC{\FFF^1}}{\XX^{KN}} &  & \\
\hline
 & \Sigma_{\ZZZ1}^{KN}{V_1^{KN}}^T &  & \\
\end{array}\right)  \\
&\FOURC
{\FPP{\XX^1}{\epsilon}}
{\FPP{\XX^{KN}}{\epsilon}}
{{V_1^1}^T\FPP{\SVEC{\ZZZ^1}}{\epsilon}}
{{V_1^{KN}}^T\FPP{\SVEC{\ZZZ^{KN}}}{\epsilon}}+
\FOURC
{-\FPP{\SVEC{\FFF^1}}{\XX^1}^TV_2^1{\Sigma_{\FFF1}^1}^{-1}{V_2^1}^Tb_2}
{b_1-\FPP{\SVEC{\FFF^1}}{\XX^{KN}}^TV_2^1{\Sigma_{\FFF1}^1}^{-1}{V_2^1}^Tb_2}
{{\Sigma_{\ZZZ1}^1}^{-1}{V_1^1}^Tb_2}
{0}=0.
\end{align*}
\end{tiny}
The main benefit of the system reduction is that the left-hand-side of this matrix becomes symmetric (after removing $\Sigma_{\ZZZ1}^1$ and $\Sigma_{\ZZZ1}^{KN}$ from the last two rows, respectively). We solve this reduced sensitivity equation using the rank-revealing $PLD(PL)^T$ factorization.
\twocolumn
\clearpage
\bibliographystyle{plainnat}
\bibliography{references}

\begin{thebibliography}{64}
\providecommand{\natexlab}[1]{#1}
\providecommand{\url}[1]{\texttt{#1}}
\expandafter\ifx\csname urlstyle\endcsname\relax
  \providecommand{\doi}[1]{doi: #1}\else
  \providecommand{\doi}{doi: \begingroup \urlstyle{rm}\Url}\fi

\bibitem[Alliez et~al.(2019)Alliez, Tayeb, and Wormser]{cgal:atw-aabb-19b}
Pierre Alliez, St{\'e}phane Tayeb, and Camille Wormser.
\newblock {3D} fast intersection and distance computation.
\newblock In \emph{{CGAL} User and Reference Manual}. {CGAL Editorial Board},
  {4.14.1} edition, 2019.
\newblock URL
  \url{https://doc.cgal.org/4.14.1/Manual/packages.html#PkgAABBTree}.

\bibitem[Amos and Kolter(2017)]{amos2017optnet}
Brandon Amos and J~Zico Kolter.
\newblock Optnet: Differentiable optimization as a layer in neural networks.
\newblock In \emph{Proceedings of the 34th International Conference on Machine
  Learning-Volume 70}, pages 136--145. JMLR. org, 2017.

\bibitem[B{\ae}rentzen(2005)]{baerentzen2005robust}
J~Andreas B{\ae}rentzen.
\newblock Robust generation of signed distance fields from triangle meshes.
\newblock In \emph{Fourth International Workshop on Volume Graphics, 2005.},
  pages 167--239. IEEE, 2005.

\bibitem[{Bicchi}(2000)]{897777}
A.~{Bicchi}.
\newblock Hands for dexterous manipulation and robust grasping: a difficult
  road toward simplicity.
\newblock \emph{IEEE Transactions on Robotics and Automation}, 16\penalty0
  (6):\penalty0 652--662, Dec 2000.
\newblock ISSN 2374-958X.
\newblock \doi{10.1109/70.897777}.

\bibitem[{Bousmalis} et~al.(2018){Bousmalis}, {Irpan}, {Wohlhart}, {Bai},
  {Kelcey}, {Kalakrishnan}, {Downs}, {Ibarz}, {Pastor}, {Konolige}, {Levine},
  and {Vanhoucke}]{8460875}
K.~{Bousmalis}, A.~{Irpan}, P.~{Wohlhart}, Y.~{Bai}, M.~{Kelcey},
  M.~{Kalakrishnan}, L.~{Downs}, J.~{Ibarz}, P.~{Pastor}, K.~{Konolige},
  S.~{Levine}, and V.~{Vanhoucke}.
\newblock Using simulation and domain adaptation to improve efficiency of deep
  robotic grasping.
\newblock In \emph{2018 IEEE International Conference on Robotics and
  Automation (ICRA)}, pages 4243--4250, May 2018.
\newblock \doi{10.1109/ICRA.2018.8460875}.

\bibitem[Brochu et~al.(2012)Brochu, Edwards, and Bridson]{brochu2012efficient}
Tyson Brochu, Essex Edwards, and Robert Bridson.
\newblock Efficient geometrically exact continuous collision detection.
\newblock \emph{ACM Transactions on Graphics (TOG)}, 31\penalty0 (4):\penalty0
  96, 2012.

\bibitem[Calli et~al.(2015)Calli, Walsman, Singh, Srinivasa, Abbeel, and
  Dollar]{calli2015benchmarking}
Berk Calli, Aaron Walsman, Arjun Singh, Siddhartha Srinivasa, Pieter Abbeel,
  and Aaron~M Dollar.
\newblock Benchmarking in manipulation research: The ycb object and model set
  and benchmarking protocols.
\newblock \emph{arXiv preprint arXiv:1502.03143}, 2015.

\bibitem[Chang et~al.(2015)Chang, Funkhouser, Guibas, Hanrahan, Huang, Li,
  Savarese, Savva, Song, Su, et~al.]{chang2015shapenet}
Angel~X Chang, Thomas Funkhouser, Leonidas Guibas, Pat Hanrahan, Qixing Huang,
  Zimo Li, Silvio Savarese, Manolis Savva, Shuran Song, Hao Su, et~al.
\newblock Shapenet: An information-rich 3d model repository.
\newblock \emph{arXiv preprint arXiv:1512.03012}, 2015.

\bibitem[Chen and Burdick(1993)]{chen1993finding}
I-Ming Chen and Joel~W Burdick.
\newblock Finding antipodal point grasps on irregularly shaped objects.
\newblock \emph{IEEE transactions on Robotics and Automation}, 9\penalty0
  (4):\penalty0 507--512, 1993.

\bibitem[{Choi} et~al.(2018){Choi}, {Schwarting}, {DelPreto}, and
  {Rus}]{8304630}
C.~{Choi}, W.~{Schwarting}, J.~{DelPreto}, and D.~{Rus}.
\newblock Learning object grasping for soft robot hands.
\newblock \emph{IEEE Robotics and Automation Letters}, 3\penalty0 (3):\penalty0
  2370--2377, July 2018.
\newblock ISSN 2377-3774.
\newblock \doi{10.1109/LRA.2018.2810544}.

\bibitem[Ciocarlie et~al.(2007)Ciocarlie, Goldfeder, and
  Allen]{ciocarlie2007dexterous}
Matei Ciocarlie, Corey Goldfeder, and Peter Allen.
\newblock Dexterous grasping via eigengrasps: A low-dimensional approach to a
  high-complexity problem.
\newblock In \emph{Robotics: Science and Systems Manipulation Workshop-Sensing
  and Adapting to the Real World}. Citeseer, 2007.

\bibitem[Community(2018)]{blender}
Blender~Online Community.
\newblock \emph{Blender - a 3D modelling and rendering package}.
\newblock Blender Foundation, Stichting Blender Foundation, Amsterdam, 2018.
\newblock URL \url{http://www.blender.org}.

\bibitem[Company(2015)]{shadowdocs}
The Shadow~Robot Company.
\newblock \emph{Shadow Robot’s Documentation}, 2015.
\newblock \url{https://shadow-robot.readthedocs.io/en/latest/index.html}.

\bibitem[Dai et~al.(2018)Dai, Majumdar, and Tedrake]{Dai2018}
Hongkai Dai, Anirudha Majumdar, and Russ Tedrake.
\newblock \emph{Synthesis and Optimization of Force Closure Grasps via
  Sequential Semidefinite Programming}, pages 285--305.
\newblock Springer International Publishing, Cham, 2018.

\bibitem[Fang et~al.(2018)Fang, Bai, Hinterstoisser, Savarese, and
  Kalakrishnan]{fang2018mtda}
Kuan Fang, Yunfei Bai, Stefan Hinterstoisser, Silvio Savarese, and Mrinal
  Kalakrishnan.
\newblock Multi-task domain adaptation for deep learning of instance grasping
  from simulation.
\newblock \emph{IEEE International Conference on Robotics and Automation
  (ICRA)}, 2018.

\bibitem[{Ferrari} and {Canny}(1992)]{219918}
C.~{Ferrari} and J.~{Canny}.
\newblock Planning optimal grasps.
\newblock In \emph{Proceedings 1992 IEEE International Conference on Robotics
  and Automation}, pages 2290--2295 vol.3, May 1992.
\newblock \doi{10.1109/ROBOT.1992.219918}.

\bibitem[Granlund and Team(2015)]{Granlund:2015:GMM:2911024}
Torbjrn Granlund and Gmp~Development Team.
\newblock \emph{GNU MP 6.0 Multiple Precision Arithmetic Library}.
\newblock Samurai Media Limited, United Kingdom, 2015.
\newblock ISBN 9789888381968, 9888381962.

\bibitem[Gualtieri et~al.(2016)Gualtieri, Ten~Pas, Saenko, and
  Platt]{gualtieri2016high}
Marcus Gualtieri, Andreas Ten~Pas, Kate Saenko, and Robert Platt.
\newblock High precision grasp pose detection in dense clutter.
\newblock In \emph{2016 IEEE/RSJ International Conference on Intelligent Robots
  and Systems (IROS)}, pages 598--605. IEEE, 2016.

\bibitem[Hang et~al.(2017)Hang, Stork, Pollard, and Kragic]{hang2017a}
K.~Hang, J.~A. Stork, N.~S. Pollard, and D.~Kragic.
\newblock A framework for optimal grasp contact planning.
\newblock \emph{IEEE Robotics and Automation Letters}, 2\penalty0 (2):\penalty0
  704--711, April 2017.
\newblock ISSN 2377-3766.
\newblock \doi{10.1109/LRA.2017.2651381}.

\bibitem[Hu et~al.(2019{\natexlab{a}})Hu, Anderson, Li, Sun, Carr,
  Ragan-Kelley, and Durand]{hu2019difftaichi}
Yuanming Hu, Luke Anderson, Tzu-Mao Li, Qi~Sun, Nathan Carr, Jonathan
  Ragan-Kelley, and Fr{\'e}do Durand.
\newblock Difftaichi: Differentiable programming for physical simulation.
\newblock \emph{arXiv preprint arXiv:1910.00935}, 2019{\natexlab{a}}.

\bibitem[Hu et~al.(2019{\natexlab{b}})Hu, Liu, Spielberg, Tenenbaum, Freeman,
  Wu, Rus, and Matusik]{hu2019chainqueen}
Yuanming Hu, Jiancheng Liu, Andrew Spielberg, Joshua~B Tenenbaum, William~T
  Freeman, Jiajun Wu, Daniela Rus, and Wojciech Matusik.
\newblock Chainqueen: A real-time differentiable physical simulator for soft
  robotics.
\newblock In \emph{2019 International Conference on Robotics and Automation
  (ICRA)}, pages 6265--6271. IEEE, 2019{\natexlab{b}}.

\bibitem[James et~al.(2019)James, Wohlhart, Kalakrishnan, Kalashnikov, Irpan,
  Ibarz, Levine, Hadsell, and Bousmalis]{james2019sim}
Stephen James, Paul Wohlhart, Mrinal Kalakrishnan, Dmitry Kalashnikov, Alex
  Irpan, Julian Ibarz, Sergey Levine, Raia Hadsell, and Konstantinos Bousmalis.
\newblock Sim-to-real via sim-to-sim: Data-efficient robotic grasping via
  randomized-to-canonical adaptation networks.
\newblock In \emph{Proceedings of the IEEE Conference on Computer Vision and
  Pattern Recognition}, pages 12627--12637, 2019.

\bibitem[{Jameson} and {Leifer}(1987)]{6499286}
J.~W. {Jameson} and L.~J. {Leifer}.
\newblock Automatic grasping: An optimization approach.
\newblock \emph{IEEE Transactions on Systems, Man, and Cybernetics},
  17\penalty0 (5):\penalty0 806--814, Sep. 1987.
\newblock ISSN 2168-2909.
\newblock \doi{10.1109/TSMC.1987.6499286}.

\bibitem[Jiang et~al.(2011)Jiang, Moseson, and Saxena]{jiang2011efficient}
Yun Jiang, Stephen Moseson, and Ashutosh Saxena.
\newblock Efficient grasping from rgbd images: Learning using a new rectangle
  representation.
\newblock In \emph{2011 IEEE International Conference on Robotics and
  Automation}, pages 3304--3311. IEEE, 2011.

\bibitem[Kappler et~al.(2015)Kappler, Bohg, and Schaal]{2015_ICRA_kbs}
D.~Kappler, B.~Bohg, and S.~Schaal.
\newblock Leveraging big data for grasp planning.
\newblock In \emph{Proceedings of the IEEE International Conference on Robotics
  and Automation}, may 2015.

\bibitem[Kasper et~al.(2012)Kasper, Xue, and Dillmann]{kasper2012kit}
Alexander Kasper, Zhixing Xue, and R{\"u}diger Dillmann.
\newblock The kit object models database: An object model database for object
  recognition, localization and manipulation in service robotics.
\newblock \emph{The International Journal of Robotics Research}, 31\penalty0
  (8):\penalty0 927--934, 2012.

\bibitem[Kiatos and Malassiotis(2019)]{kiatos2019grasping}
Marios Kiatos and Sotiris Malassiotis.
\newblock Grasping unknown objects by exploiting complementarity with robot
  hand geometry.
\newblock In \emph{International Conference on Computer Vision Systems}, pages
  88--97. Springer, 2019.

\bibitem[Kingma and Ba(2015)]{kingma2014adam}
Diederik~P Kingma and Jimmy Ba.
\newblock Adam: A method for stochastic optimization.
\newblock In \emph{Proceedings of International Conference on Learning
  Representations}, 2015.

\bibitem[Liu et~al.(2019{\natexlab{a}})Liu, Pan, Xu, Ganguly, and
  Manocha]{liu2019grasp}
Min Liu, Zherong Pan, Kai Xu, Kanishka Ganguly, and Dinesh Manocha.
\newblock Generating grasp poses for a high-dof gripper using neural networks.
\newblock In \emph{Proceedings of 2019 IEEE/RSJ International Conference on
  Intelligent Robots and Systems}, 2019{\natexlab{a}}.

\bibitem[Liu et~al.(2019{\natexlab{b}})Liu, Pan, Xu, and Manocha]{1909.05430}
Min Liu, Zherong Pan, Kai Xu, and Dinesh Manocha.
\newblock New formulation of mixed-integer conic programming for globally
  optimal grasp planning.
\newblock In \emph{arXiv:1909.05430v3}, 2019{\natexlab{b}}.

\bibitem[Lu et~al.(2020{\natexlab{a}})Lu, Chenna, Sundaralingam, and
  Hermans]{lu2020planning}
Qingkai Lu, Kautilya Chenna, Balakumar Sundaralingam, and Tucker Hermans.
\newblock Planning multi-fingered grasps as probabilistic inference in a
  learned deep network.
\newblock In \emph{Robotics Research}, pages 455--472. Springer,
  2020{\natexlab{a}}.

\bibitem[Lu et~al.(2020{\natexlab{b}})Lu, Van~der Merwe, Sundaralingam, and
  Hermans]{lu2020multifingered}
Qingkai Lu, Mark Van~der Merwe, Balakumar Sundaralingam, and Tucker Hermans.
\newblock Multifingered grasp planning via inference in deep neural networks:
  Outperforming sampling by learning differentiable models.
\newblock \emph{IEEE Robotics \& Automation Magazine}, 2020{\natexlab{b}}.

\bibitem[Mahler et~al.(2016)Mahler, Pokorny, Hou, Roderick, Laskey, Aubry,
  Kohlhoff, Kr{\"o}ger, Kuffner, and Goldberg]{mahler2016dex}
Jeffrey Mahler, Florian~T Pokorny, Brian Hou, Melrose Roderick, Michael Laskey,
  Mathieu Aubry, Kai Kohlhoff, Torsten Kr{\"o}ger, James Kuffner, and Ken
  Goldberg.
\newblock Dex-net 1.0: A cloud-based network of 3d objects for robust grasp
  planning using a multi-armed bandit model with correlated rewards.
\newblock In \emph{2016 IEEE International Conference on Robotics and
  Automation (ICRA)}, pages 1957--1964. IEEE, 2016.

\bibitem[Mahler et~al.(2017)Mahler, Liang, Niyaz, Laskey, Doan, Liu, Aparicio,
  and Goldberg]{inproceedingsDexNetTwo}
Jeffrey Mahler, Jacky Liang, Sherdil Niyaz, Michael Laskey, Richard Doan, Xinyu
  Liu, Juan Aparicio, and Ken Goldberg.
\newblock Dex-net 2.0: Deep learning to plan robust grasps with synthetic point
  clouds and analytic grasp metrics.
\newblock In \emph{Proceedings of Robotics: Science and Systems}, 07 2017.
\newblock \doi{10.15607/RSS.2017.XIII.058}.

\bibitem[Maldonado et~al.(2010)Maldonado, Klank, and
  Beetz]{maldonado2010robotic}
Alexis Maldonado, Ulrich Klank, and Michael Beetz.
\newblock Robotic grasping of unmodeled objects using time-of-flight range data
  and finger torque information.
\newblock In \emph{2010 IEEE/RSJ International Conference on Intelligent Robots
  and Systems}, pages 2586--2591. IEEE, 2010.

\bibitem[Miller and Allen(2000)]{miller2000graspit}
Andrew~T Miller and Peter~K Allen.
\newblock Graspit!: A versatile simulator for grasp analysis.
\newblock In \emph{in Proc. of the ASME Dynamic Systems and Control Division}.
  Citeseer, 2000.

\bibitem[Miller(1997)]{miller1997sensitivity}
Scott~A Miller.
\newblock Sensitivity of solutions to semidefinite programs.
\newblock 1997.

\bibitem[Mordatch et~al.(2012)Mordatch, Todorov, and
  Popovi{\'c}]{mordatch2012discovery}
Igor Mordatch, Emanuel Todorov, and Zoran Popovi{\'c}.
\newblock Discovery of complex behaviors through contact-invariant
  optimization.
\newblock \emph{ACM Transactions on Graphics (TOG)}, 31\penalty0 (4):\penalty0
  1--8, 2012.

\bibitem[Morrison et~al.(2018)Morrison, Corke, and
  Leitner]{morrison2018closing}
Douglas Morrison, Peter Corke, and J{\"u}rgen Leitner.
\newblock Closing the loop for robotic grasping: A real-time, generative grasp
  synthesis approach.
\newblock In \emph{2018 Robotics: Science and Systems}, 2018.

\bibitem[Mousavian et~al.(2019)Mousavian, Eppner, and Fox]{Mousavian_2019_ICCV}
Arsalan Mousavian, Clemens Eppner, and Dieter Fox.
\newblock 6-dof graspnet: Variational grasp generation for object manipulation.
\newblock In \emph{The IEEE International Conference on Computer Vision
  (ICCV)}, October 2019.

\bibitem[Paszke et~al.(2017)Paszke, Gross, Chintala, Chanan, Yang, DeVito, Lin,
  Desmaison, Antiga, and Lerer]{paszke2017automatic}
Adam Paszke, Sam Gross, Soumith Chintala, Gregory Chanan, Edward Yang, Zachary
  DeVito, Zeming Lin, Alban Desmaison, Luca Antiga, and Adam Lerer.
\newblock Automatic differentiation in {PyTorch}.
\newblock In \emph{NIPS Autodiff Workshop}, 2017.

\bibitem[Quillen et~al.(2018)Quillen, Jang, Nachum, Finn, Ibarz, and
  Levine]{quillen2018deep}
Deirdre Quillen, Eric Jang, Ofir Nachum, Chelsea Finn, Julian Ibarz, and Sergey
  Levine.
\newblock Deep reinforcement learning for vision-based robotic grasping: A
  simulated comparative evaluation of off-policy methods.
\newblock In \emph{2018 IEEE International Conference on Robotics and
  Automation (ICRA)}, pages 6284--6291. IEEE, 2018.

\bibitem[Ravi et~al.(2019)Ravi, Dinh, Lokhande, and Singh]{ravi2019explicitly}
Sathya~N Ravi, Tuan Dinh, Vishnu~Suresh Lokhande, and Vikas Singh.
\newblock Explicitly imposing constraints in deep networks via conditional
  gradients gives improved generalization and faster convergence.
\newblock In \emph{Proceedings of the AAAI Conference on Artificial
  Intelligence}, volume~33, pages 4772--4779, 2019.

\bibitem[Roa and Su{\'a}rez(2015)]{roa2015grasp}
M{\'a}ximo~A Roa and Ra{\'u}l Su{\'a}rez.
\newblock Grasp quality measures: review and performance.
\newblock \emph{Autonomous robots}, 38\penalty0 (1):\penalty0 65--88, 2015.

\bibitem[Saxena et~al.(2007)Saxena, Driemeyer, Kearns, and
  Ng]{saxena2007robotic}
Ashutosh Saxena, Justin Driemeyer, Justin Kearns, and Andrew~Y Ng.
\newblock Robotic grasping of novel objects.
\newblock In \emph{Advances in neural information processing systems}, pages
  1209--1216, 2007.

\bibitem[Saxena et~al.(2008)Saxena, Driemeyer, and Ng]{saxena2008robotic}
Ashutosh Saxena, Justin Driemeyer, and Andrew~Y Ng.
\newblock Robotic grasping of novel objects using vision.
\newblock \emph{The International Journal of Robotics Research}, 27\penalty0
  (2):\penalty0 157--173, 2008.

\bibitem[{Schmidt} et~al.(2018){Schmidt}, {Vahrenkamp}, {Wächter}, and
  {Asfour}]{8463204}
P.~{Schmidt}, N.~{Vahrenkamp}, M.~{Wächter}, and T.~{Asfour}.
\newblock Grasping of unknown objects using deep convolutional neural networks
  based on depth images.
\newblock In \emph{2018 IEEE International Conference on Robotics and
  Automation (ICRA)}, pages 6831--6838, May 2018.
\newblock \doi{10.1109/ICRA.2018.8463204}.

\bibitem[Schulman et~al.(2017)Schulman, Goldberg, and
  Abbeel]{schulman2017grasping}
John~D Schulman, Ken Goldberg, and Pieter Abbeel.
\newblock Grasping and fixturing as submodular coverage problems.
\newblock In \emph{Robotics Research}, pages 571--583. Springer, 2017.

\bibitem[Singh et~al.(2014)Singh, Sha, Narayan, Achim, and
  Abbeel]{singh2014bigbird}
Arjun Singh, James Sha, Karthik~S Narayan, Tudor Achim, and Pieter Abbeel.
\newblock Bigbird: A large-scale 3d database of object instances.
\newblock In \emph{2014 IEEE International Conference on Robotics and
  Automation (ICRA)}, pages 509--516. IEEE, 2014.

\bibitem[Su et~al.(2015)Su, Maji, Kalogerakis, and Learned-Miller]{su2015multi}
Hang Su, Subhransu Maji, Evangelos Kalogerakis, and Erik Learned-Miller.
\newblock Multi-view convolutional neural networks for 3d shape recognition.
\newblock In \emph{Proceedings of the IEEE international conference on computer
  vision}, pages 945--953, 2015.

\bibitem[Tassa et~al.(2012)Tassa, Erez, and Todorov]{tassa2012synthesis}
Yuval Tassa, Tom Erez, and Emanuel Todorov.
\newblock Synthesis and stabilization of complex behaviors through online
  trajectory optimization.
\newblock In \emph{2012 IEEE/RSJ International Conference on Intelligent Robots
  and Systems}, pages 4906--4913. IEEE, 2012.

\bibitem[Tobin et~al.(2017)Tobin, Fong, Ray, Schneider, Zaremba, and
  Abbeel]{tobin2017domain}
Josh Tobin, Rachel Fong, Alex Ray, Jonas Schneider, Wojciech Zaremba, and
  Pieter Abbeel.
\newblock Domain randomization for transferring deep neural networks from
  simulation to the real world.
\newblock In \emph{2017 IEEE/RSJ International Conference on Intelligent Robots
  and Systems (IROS)}, pages 23--30. IEEE, 2017.

\bibitem[Van~der Merwe et~al.(2019)Van~der Merwe, Lu, Sundaralingam, Matak, and
  Hermans]{van2019learning}
Mark Van~der Merwe, Qingkai Lu, Balakumar Sundaralingam, Martin Matak, and
  Tucker Hermans.
\newblock Learning continuous 3d reconstructions for geometrically aware
  grasping.
\newblock \emph{arXiv preprint arXiv:1910.00983}, 2019.

\bibitem[Villegas et~al.(2018)Villegas, Yang, Ceylan, and
  Lee]{villegas2018neural}
Ruben Villegas, Jimei Yang, Duygu Ceylan, and Honglak Lee.
\newblock Neural kinematic networks for unsupervised motion retargetting.
\newblock In \emph{Proceedings of the IEEE Conference on Computer Vision and
  Pattern Recognition}, pages 8639--8648, 2018.

\bibitem[Wang and Hauser(2019{\natexlab{a}})]{Hauser-RSS-19}
Fan Wang and Kris Hauser.
\newblock Robot packing with known items and nondeterministic arrival order.
\newblock In \emph{Proceedings of Robotics: Science and Systems},
  FreiburgimBreisgau, Germany, June 2019{\natexlab{a}}.
\newblock \doi{10.15607/RSS.2019.XV.035}.

\bibitem[Wang and Hauser(2019{\natexlab{b}})]{wang2019stable}
Fan Wang and Kris Hauser.
\newblock Stable bin packing of non-convex 3d objects with a robot manipulator.
\newblock In \emph{2019 International Conference on Robotics and Automation
  (ICRA)}, pages 8698--8704. IEEE, 2019{\natexlab{b}}.

\bibitem[{Yan} et~al.(2018){Yan}, {Hsu}, {Khansari}, {Bai}, {Pathak}, {Gupta},
  {Davidson}, and {Lee}]{8460609}
X.~{Yan}, J.~{Hsu}, M.~{Khansari}, Y.~{Bai}, A.~{Pathak}, A.~{Gupta},
  J.~{Davidson}, and H.~{Lee}.
\newblock Learning 6-dof grasping interaction via deep geometry-aware 3d
  representations.
\newblock In \emph{2018 IEEE International Conference on Robotics and
  Automation (ICRA)}, pages 3766--3773, May 2018.
\newblock \doi{10.1109/ICRA.2018.8460609}.

\bibitem[{Zheng}(2013)]{6335488}
Y.~{Zheng}.
\newblock An efficient algorithm for a grasp quality measure.
\newblock \emph{IEEE Transactions on Robotics}, 29\penalty0 (2):\penalty0
  579--585, April 2013.
\newblock ISSN 1941-0468.
\newblock \doi{10.1109/TRO.2012.2222274}.

\bibitem[{Zheng}(2016)]{7539662}
Y.~{Zheng}.
\newblock Computing the globally optimal frictionless fixture in a discrete
  point set.
\newblock \emph{IEEE Transactions on Robotics}, 32\penalty0 (4):\penalty0
  1026--1032, Aug 2016.
\newblock ISSN 1941-0468.
\newblock \doi{10.1109/TRO.2016.2588720}.

\bibitem[{Zheng}(2017)]{7989253}
Y.~{Zheng}.
\newblock Computing the best grasp in a discrete point set.
\newblock In \emph{2017 IEEE International Conference on Robotics and
  Automation (ICRA)}, pages 2208--2214, May 2017.
\newblock \doi{10.1109/ICRA.2017.7989253}.

\bibitem[Zheng(2012)]{zheng2012efficient}
Yu~Zheng.
\newblock An efficient algorithm for a grasp quality measure.
\newblock \emph{IEEE Transactions on Robotics}, 29\penalty0 (2):\penalty0
  579--585, 2012.

\bibitem[Zhou et~al.(2019)Zhou, Hou, and Mason]{doi:10.1177/0278364919872532}
Jiaji Zhou, Yifan Hou, and Matthew~T Mason.
\newblock Pushing revisited: Differential flatness, trajectory planning, and
  stabilization.
\newblock \emph{The International Journal of Robotics Research}, 38\penalty0
  (12-13):\penalty0 1477--1489, 2019.
\newblock \doi{10.1177/0278364919872532}.

\bibitem[Zhou and Jacobson(2016)]{zhou2016thingi10k}
Qingnan Zhou and Alec Jacobson.
\newblock Thingi10k: A dataset of 10,000 3d-printing models.
\newblock \emph{arXiv preprint arXiv:1605.04797}, 2016.

\bibitem[Zhu et~al.(2019)Zhu, Gupta, Rajeswaran, Levine, and
  Kumar]{zhu2019dexterous}
Henry Zhu, Abhishek Gupta, Aravind Rajeswaran, Sergey Levine, and Vikash Kumar.
\newblock Dexterous manipulation with deep reinforcement learning: Efficient,
  general, and low-cost.
\newblock In \emph{2019 International Conference on Robotics and Automation
  (ICRA)}, pages 3651--3657. IEEE, 2019.

\end{thebibliography}
\end{document}